\documentclass[final]{cvpr}

\usepackage{times}
\usepackage{epsfig}
\usepackage{graphicx}
\usepackage{amsmath}
\usepackage{amssymb}

\usepackage{booktabs} % To thicken table lines
\usepackage{soul,color}
\usepackage{nicefrac}
\usepackage{multirow}
\usepackage{sidecap}
\usepackage{amsmath}

\usepackage{footnote}
\makesavenoteenv{tabular}
\makesavenoteenv{table}

\usepackage{caption}
\usepackage{tablefootnote}

\usepackage{balance}

\usepackage[bottom]{footmisc}

\usepackage{xcolor}
\definecolor{lightgreen}{rgb}{0.20,0.67,0.25}

\usepackage[pagebackref=true,breaklinks=true,colorlinks,bookmarks=false]{hyperref}

\def\cvprPaperID{4435} % *** Enter the CVPR Paper ID here
\def\confYear{CVPR 2021}

\begin{document}

%%%%%%%%% TITLE
\title{Learning from Weakly-labeled Web Videos via Exploring Sub-Concepts}

% \and

\author{% Kunpeng, zizhao, guanhang, xuehan, chenyu, fuyun, tomas
   Kunpeng Li$^{1}$\thanks{Work done while the author was an intern at Google.} \quad Zizhao Zhang$^{2}$ \quad Guanhang Wu$^{2}$ \quad Xuehan Xiong$^{2}$ \quad 
    Chen-Yu Lee$^{2}$ \\ \quad Zhichao Lu$^{2}$ \quad Yun Fu$^{1}$ \quad Tomas Pfister$^{2}$ \\
   $^1$Northeastern University \quad  $^2$Google Cloud AI\\
}

\maketitle

%%%%%%%%% ABSTRACT
\begin{abstract}
Learning visual knowledge from massive weakly-labeled web videos has attracted growing research interests thanks to the large corpus of easily accessible video data on the Internet. However, for video action recognition, the action of interest might only exist in arbitrary clips of untrimmed web videos, resulting in high label noises in the temporal space. To address this issue, we introduce a new method for pre-training video action recognition models using queried web videos.
Instead of trying to filter out, we propose to convert the potential noises in these queried videos to useful supervision signals by defining the concept of Sub-Pseudo Label (SPL).
Specifically, SPL spans out a new set of meaningful ``middle ground'' label space constructed by extrapolating the original weak labels during video querying and the prior knowledge distilled from a teacher model. Consequently, SPL provides enriched supervision for video models to learn better representations. SPL is fairly simple and orthogonal to popular teacher-student self-training frameworks without extra training cost.
We validate the effectiveness of our method on four video action recognition datasets and a weakly-labeled image dataset to study the generalization ability.
Experiments show that SPL outperforms several existing pre-training strategies using pseudo-labels and the learned representations lead to competitive results when fine-tuning on HMDB-51 and UCF-101 compared with recent pre-training methods.
\end{abstract}

%%%%%%%%% BODY TEXT
\section{Introduction}

Remarkable successes \cite{simonyan2014two,tran2015learning,feichtenhofer2019slowfast} have been achieved in video recognition in recent years thanks to the development of deep learning models. 
However, training deep neural networks requires a large amount of 
human-annotated data. It requires tremendous human labor and huge financial cost and therefore oftentimes sets out to be the bottleneck for real-world video recognition applications.

\begin{figure}
\centering
\includegraphics[width=0.88\linewidth]{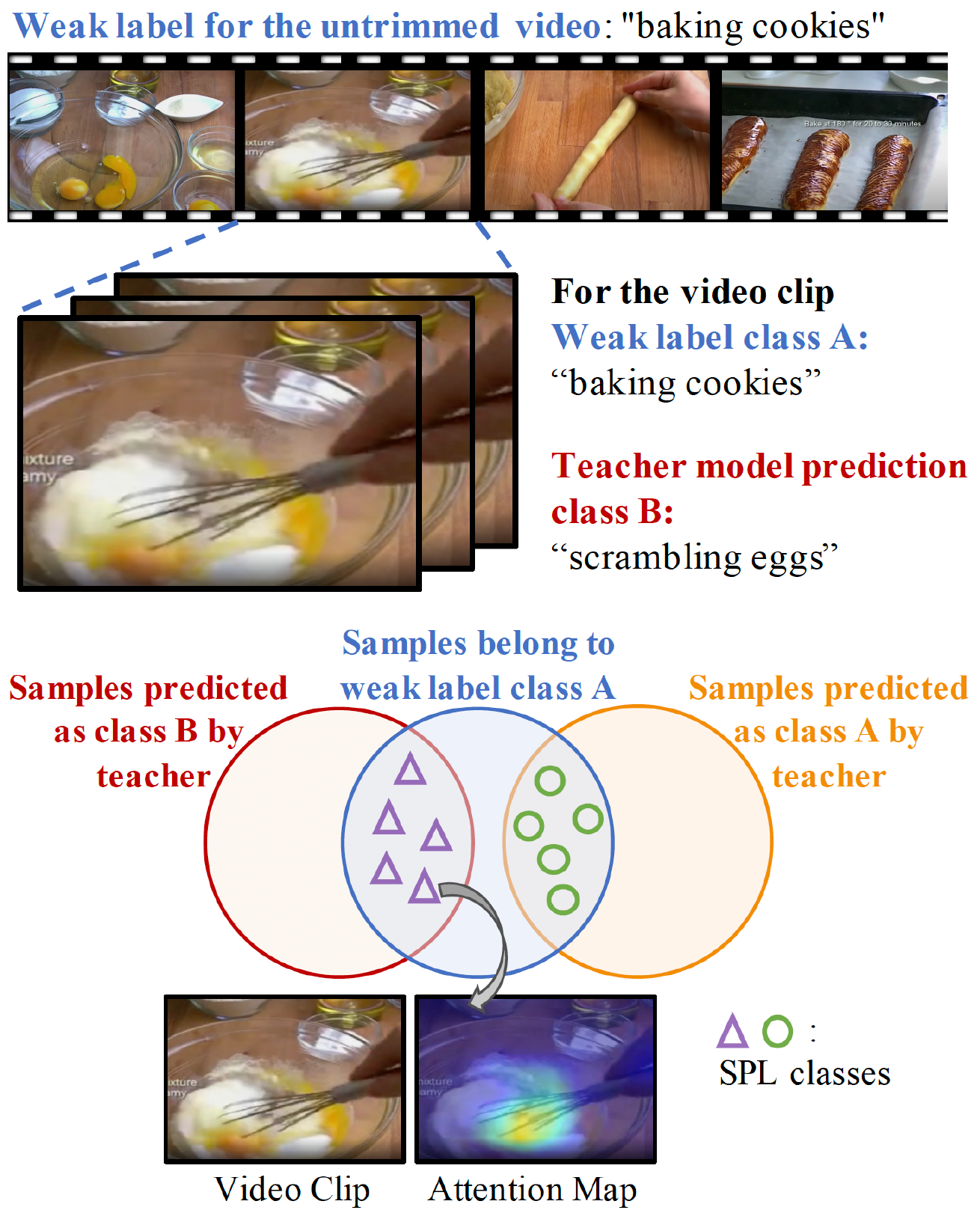} 
\vspace{-3mm}
\caption{Illustration of the motivation. Sampled training clips may represent a different visual action (scrambling eggs) from the query label of the whole untrimmed video (backing cookies). 
SPL converts the potential noises to useful supervision signals by creating a new set of meaningful ``middle ground'' pseudo labels (i.e. sub-concepts) via extrapolating two related action classes. Enriched supervision is provided for effective model pre-training.} %Overview of SPL. Both weak label A and model prediction B are conceptually correct classes. We create SPLs to discover new concepts for effective video model pre-training.
\label{fig:teaser}
\vspace{-5mm}
\end{figure}

Web videos are usually acquired by online querying through label keywords. A keyword, which we refer as a weak label, is then assigned to each untrimmed video obtained.
Although large-scale videos with weak labels are easier to be collected, training with un-verified weak labels poses another challenge in developing robust models. 
%There are rapid growing research interests in exploring effective strategies of learning from weakly-labeled web videos \cite{ghadiyaram2019large,yalniz2019billion}.
%Therefore, the label noise challenges in video are multi-folds \cite{ghadiyaram2019large}.
Recent studies \cite{ghadiyaram2019large,kuehne2019mining,chang2019d3tw} have demonstrated that, in addition to the label noise (e.g. incorrect action labels on untrimmed videos), there are temporal noise due to the lack of accurate temporal action localization. 
This means an untrimmed web video may include other non-targeted content or may only contain a small proportion of the target action. Reduce noise effects for large-scale weakly-supervised pre-training is critical but particularly challenging in practice. \cite{ghadiyaram2019large} suggests to query short videos (e.g., within 1 minute) to obtain more accurate temporal localization of target actions. However, such data pre-processing method prevents models from fully utilizing widely available web video data, especially longer videos with richer contents.

% large-scale
% In this work, we propose a novel learning method to improve the web-supervised pre-training for video action recognition . 
In this work, we propose a novel learning method to conduct effective pre-training on untrimmed videos from the web.
Instead of simply filtering the potential temporal noise, we propose to convert such ``noisy'' data to useful supervision by leveraging the proposed concept of Sub-Pseudo Label (SPL). 
As shown in Figure \ref{fig:teaser}, SPL creates a new set of meaningful ``middle ground'' pseudo-labels to expand the original weak label space.
Our motivation is based on the observation that, within the same untrimmed video, these ``noisy'' video clips have semantic relations with the target action (weak label class), but may also include essential visual components of other actions (such as the teacher model predicted class). %Therefore, natural confusion exists between these target actions. 
Our method aims to use the extrapolated SPLs from weak labels and distilled labels to capture the enriched supervision signals, encouraging learning better representations during pre-training that can be used for downstream fine-tuning tasks.

Discovering meaningful SPLs has critical impact on learning high quality representations. To this end, we take advantage of the original weak labels as well as the prior knowledge distilled from a set of target labeled data from human annotations. 
Specifically, we first train a teacher model from the target labeled data and perform inference on every clip of web videos. 
From the teacher model predicted labels and the original weak labels of the web video, we design a mapping function to construct SPLs for these video clips. We then perform large-scale pre-training on web videos utilizing SPLs as the supervision space. In addition, we study different variants of mapping functions to tackle high-dimensional label space when the number of classes is high.
We construct weakly-labeled web video data based on two video datasets: Kinetics-200 \cite{xie2018rethinking,carreira2017quo} and SoccerNet \cite{giancola2018soccernet}. Experimental results show that our method can consistently improve the performance of conventional supervised methods and several existing pre-training strategies on these two datasets. 
We also follow recent works~\cite{miech2020end,patrick2020multi,stroud2020learning} to conduct experiments on HMDB-51~\cite{kuehne2011hmdb} and UCF-101~\cite{soomro2012ucf101} to show that learned representations by SPL are generic and useful when SPL pre-training is conducted on web data from different domains. 
SPL achieves competitive results on these datasets compared with recent state-of-the-art pre-training methods.

In summary, our contributions can be concluded as follows: 
(a) We propose a novel concept of SPL to provide valid supervision for learning from weakly-labeled untrimmed web videos so that better representations can be obtained after the pre-training. 
(b) We investigate several space-reduced SPL classes discovering strategies utilizing weak labels as well as the knowledge distilled from the teacher model trained on the labeled dataset. 
(c) Comprehensive experiments on multiple video datasets show that our method can consistently improve the performance of baselines on both common and fine-grained action recognition datasets. We also validate the generalization ability of the proposed method on a weakly-labeled image classification benchmark (the source code is provided). (d) SPL is fairly simple and orthogonal to popular teacher-student self-training frameworks to reduce label noises of either pseudo labels or data weak labels.

% \subsection{Style}

\vspace{-2mm}
\section{Related Work}
\vspace{-2mm}

\textbf{Learning from the web data.} There are growing studies using information from Web that aim to reduce the data annotation and collection costs \cite{chen2013neil,divvala2014learning,lee2018cleannet,mahajan2018exploring,kuehne2019mining,yan2019clusterfit,duan2020omni,yalniz2019billion}. For video classification, early works \cite{ma2017less,gan2016webly,sun2015temporal} focus on utilizing web action images to boost action recognition models. \cite{sun2015temporal} and \cite{duan2012exploiting} propose to apply domain transfer approaches to learn from Web images for event recognition and action localization. 
However these image-based methods do not consider the rich temporal dynamics of videos. 
Interactions between Web videos and images are further studied in \cite{gan2016you}, where a CNN network is first trained on web videos and further refined in a curriculum learning manner using web images. 
Following this line, \cite{rupprecht2018learning} explores using information from two independent CNNs, a RGB network trained on web videos and images as well as a second network using temporal information from optical flow. It enriches the training set with heterogeneous web data sources and validates the benefits.
Different from relying on multiple modalities, \cite{yeung2017learning} proposes a reinforcement learning-based formulation to select reliable training examples from noisy web search results. A data labeling policy is learned via Q-learning on a small labeled training set, which is then used to label noisy web data to conduct further training.
More recently, \cite{ghadiyaram2019large} demonstrates that better pre-training models can be obtained by learning from very large scale noisy web videos with short length e.g., within 1 minute.  
Differently, SPL handles the temporal noise in untrimmed videos by exploring valid sub-concepts, so that enriched supervision can be provided for effective pre-training.

\begin{figure*}
\centering
\includegraphics[width=0.75\linewidth]{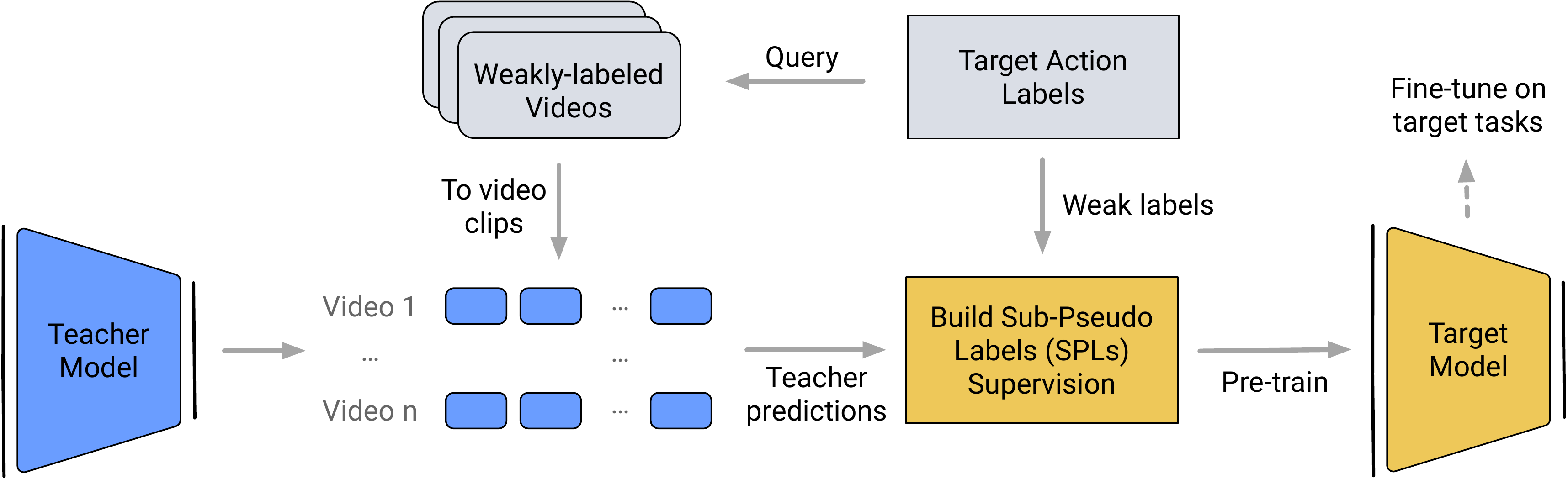}
\caption{The overall pre-training framework for learning form web videos via exploring SPLs.}
\label{fig:framework} 
\vspace{-.3cm}
\end{figure*} 
\textbf{}

\vspace{-.3cm}
\textbf{Knowledge distillation.}
Our work is also related to Knowledge Distillation 
\cite{bucilu2006model,ba2014deep,hinton2015distilling,furlanello2018born,xie2019self,yan2019clusterfit,muller2020subclass} 
% \cite{bucilu2006model,hinton2015distilling}
where the goal is to transfer knowledge from one model (the teacher) to another (the student). 
% \cite{xie2019self} shows that student models can outperform the teacher model based on unlabeled data and data augmentation.
Several recent distillation methods \cite{furlanello2018born,xie2019self} focus on teachers and students with the same architecture which can be trained sequentially for image classification. \cite{xie2019self} shows that student model can outperform the teacher model based on unlabeled data and data augmentation. 
\cite{yan2019clusterfit} finds that training student models using cluster assignments of features extracted from a teacher model can improve the generalization of visual representations.  
\cite{muller2020subclass} proposes to improve the transfer by forcing the teacher to expand the network output logits during the supervised training. 
% forcing the teacher to divide each class into many sub-classes that it invents during the supervised training. 
The student is then trained to match the subclass probabilities.
While to some extend SPL and it share similar high-level motivation of exploring sub-concepts, the differences are significant.
Methods are different: our method uses the extrapolated SPLs from weak labels and distilled labels, while \cite{muller2020subclass} learns to expend teacher network logits, which relies on a pre-defined hyper-parameter of subclass numbers.
Problems are different: \cite{muller2020subclass} focuses on the knowledge distillation between large and small networks for image classification.
% Our method shares similar motivation of discovering sub-concepts to provide useful supervision.
% But different from learns to expend teacher network logits, which relies on a pre-defined hyper-parameter of subclass numbers
% But different from them only focusing on the knowledge distilled from a teacher model, we bridge the distilled knowledge and weak labels to discover meaningful new concepts via SPL.   

%Different from existing works, we focus on the video classification tasks by proposing a new set of meaningful SPLs by taking advantage of the weak labels and the prior knowledge distilled from a teacher model.

% Several recent distillation methods \cite{furlanello2018born,xie2019self} focus on teachers and students with the same architecture which can be trained sequentially for image classification. Xie et al \cite{xie2019self} show that student model can outperform the teacher model based on unlabeled data and data augmentation. 

% Video analysis has long been tackled using hand-crafted features \cite{laptev2005space,wang2013action}. Following the success of deep learning in the image domain \cite{krizhevsky2012imagenet,he2016deep},

% Weakly labeled action recognition, such as weakly video segmentation stc. 

\textbf{Video recognition.}
 Hand-crafted features have long been used to tackle video analysis \cite{laptev2005space,wang2013action}.
%  Video analysis has long been tackled using hand-crafted features \cite{laptev2005space,wang2013action}. 
 Following the great progress of deep learning in the image domain \cite{krizhevsky2012imagenet,he2016deep}, recent video action recognition methods usually use two-stream (optical flow and RGB) networks \cite{simonyan2014two,feichtenhofer2016convolutional} or 3D ConvNets \cite{carreira2017quo,tran2015learning,feichtenhofer2019slowfast}. 
 The former one extracts image-level features using 2D networks and then conducts temporal aggregation on top, while the latter one directly learns spatial-temporal features from consecutive video frames.
%  The former one first uses 2D networks to extract image-level feature and then performs temporal aggregation on top while the latter one learns spatial-temporal features directly from video clips in the form of consecutive frames. 
 % Our work improves the performance of the action recognition model utilizing web videos.
 Our work aims to improve the performance of action recognition models by utilizing weakly-labeled web videos.

% \vspace{-3mm}
\section{Method}
% \vspace{-2mm}

\subsection{Problem formulation and method overview}\label{sc:method_overview}

%\zizhao{MixUp paper https://arxiv.org/pdf/1710.09412.pdf is a very nice example how to write a simple idea in a beautiful way}
% \subsection{Overview of the Method}

%For pre-training on a data set \(D\) with annotations
For pre-training on a dataset \(D_p\) with \(N\) target actions, we aim to learn representations that can benefit the downstream task by afterwards fine-tuning on the target dataset \(D_t\). This pre-training process of model \(M\) is usually achieved by minimizing the cross-entropy loss between the data samples \(x\) and their corresponding labels \(y\), as follows:
\begin{equation}
\label{eq:cross_entropy}
L_{\text{CE}} = -{\mathbb{E}_{(x,y) \sim{D_p}}}\sum_{c=1}^N y_c\log M(x),
\end{equation}
where \(y_c \in \{0,1\} \) indicates whether \(x\) belongs to class \(c \in [0, N-1]\).

% L_{\text{CE}} =  - {\mathbb{E}_{(x,y) \sim{D_p}}}\sum\limits_{c = 1}^N {{y_c}\log } M(x),

In the case of pre-training on a web video set as \(D_p\), we sample clips from these untrimmed web videos to construct the training data. Since there are no ground-truth annotations, assigning a valid label \(y\) for each clip sample \(x\) is a key. A common practice \cite{ghadiyaram2019large,mahajan2018exploring} is to treat the text query or hash tags that come together with the web videos as weak labels \(l\). However this causes high label and temporal noises as target actions might exist in arbitrary clips of the entire video that occupy a very small portion. In addition to relying on the weak labels, we can also distill knowledge from a teacher model \(T\) trained on the target dataset \(D_t\) using Eq. \ref{eq:cross_entropy}, where \(D_p\) is replaced by \(D_t\). A basic teacher-student training pipeline \cite{furlanello2018born,xie2019self} can be applied by treating the teacher model prediction as the pseudo-label to train a student model on \(D_p\). But there will be information lost as the original informative weak labels are totally ignored. Another strategy is to use agreement filtering to select reliable data whose weak labels match their teacher model predictions. However, in practice we find this strategy will discard a large amount of training data from \(D_p\) (over \(60\%\) in our experiments on the Kinetics-200 dataset), which limits the data scale for training deep neural networks. %These three strategies are shown in bottom-right part of Figure \ref{fig:spl_assign_vis}.  

Instead of treating the potential noise in \(D_p\) as useless data to filter out, we propose to migrate such noisy data to useful supervision by defining the concept of Sub-Pseudo Label (SPL). Specifically, SPL creates a new set of meaningful ``middle ground'' pseudo-labels, which are discovered by taking advantage of the original weak labels and the prior knowledge distilled from the teacher model. Figure \ref{fig:framework} illustrates the overall framework to utilize SPLs. %We describe more details about assigning SPL in Section \ref{sc:assign_pseudo_sublabel}. 

%Furthermore, to solve the issue due to \(O(N^2)\) label space, we further introduce several variations of our method in \ref{sc:variations}.    

\subsection{SPL for individual training sample}
\label{sc:assign_pseudo_sublabel}

To determine the SPL class for each video clip in \(D_p\), we first perform inference on each video clip in \(D_p\) using the teacher model \(T\) trained on \(D_t\). A 2-dimensional confusion matrix \(C \in \mathbb{R}^{N\times N}\) can be obtained to summarize the alignments between the teacher model inferences (columns) and the original weak annotations (rows). %The confusion matrix \(C\) with dimension of \(N \times N\), where \(N\) is the number of action classes, is not used to evaluate the performance of the teacher model but to provide information for extra valid supervision. 

Specifically, video clips at the diagonal location \((w,w)\) of \(C\) can be roughly treated as samples belonging to class \(w\), which is agreed by the original weak label as well as the teacher model \(T\). For other samples at off-diagonal location \((h,w)\) of \(C\), we interpret them as follows: from the view of the weak labels, these clips come from videos retrieved using text query of the action class \(h\). Therefore, they include context information that may not exactly represent the action class \(h\) but is semantically related to it. However, from the view of the teacher model \(T\), visual features that belong to action class \(w\) can also be found in these clips based on knowledge learned from the target dataset \(D_t\). Instead of allocating these samples to ether action class \(h\) or \(w\) with the risk of leading to label noise, we convert such confusion to a useful supervision signal by assigning SPLs. For each data sample \((x, y)\) in \(D_p\), the sub-pseudo label \(y \in [0, N^2-1]\) of the video clip \(x\) is obtained by
% \({p_i} = h + w\), where (h,w) is location of confusion matrix \(C\) that the current video clip \(i\) belongs to. 
% \zizhao{Maybe no need to use location h and w to refer labels. You can just define two labels: weak labels $y$ and teacher predictions $T(x)$ and do $SPL(x) = y_x + T(x)$}
\begin{equation}
\label{eq: sub_pseudo_label}
y = N\cdot l + T(x),
% {p_i} = h + w,
\end{equation} 
where \(l\) is the weak label of \(x\) and \(T(x)\) is its teacher prediction, where \(l, T(x) \in [0, N-1]\).
% where (h,w) is location of confusion matrix \(C\) that the current video clip \(i\) belongs to.

\subsection{Reduction of the quadratic SPL space}\label{sc:variations}

Given \(N\) categories of original labels, SPL results in a quadratic label space \(O(N^2)\). When \(N\) is big, the softmax output layer becomes too large to train 3D video models efficiently. Moreover, the distribution of SPL classes is usually long-tailed as some semantically distant classes can be unlikely confused with each other. %For example, surfing and playing basketball can have very distinct visual content. 
%Hence forcing the model to learn information of imbalanced tail SPL classes can be less effective. Instead, 
We believe head SPL classes contain more useful supervision information than tails. Based on this motivation, we propose several variations of SPLs to reduce the SPL space. 
% illustrates each variants of SPL of the followings.

\textbf{Merge to Diagonal (SPL-M)}: Suppose we are targeting at SPLs with a total number of \(K \times N\) classes, we kept \(N\) in-diagonal classes and then select the most frequent \( (K-1) \times N\) off-diagonal classes as new SPLs. For the samples of un-selected off-diagonal classes, we merge them into the diagonals of their corresponding rows. Since each row belongs to a class of weak labels, this strategy promotes original weak labels over teacher predictions.

\begin{figure}
\centering
\includegraphics[width=1.0\linewidth]{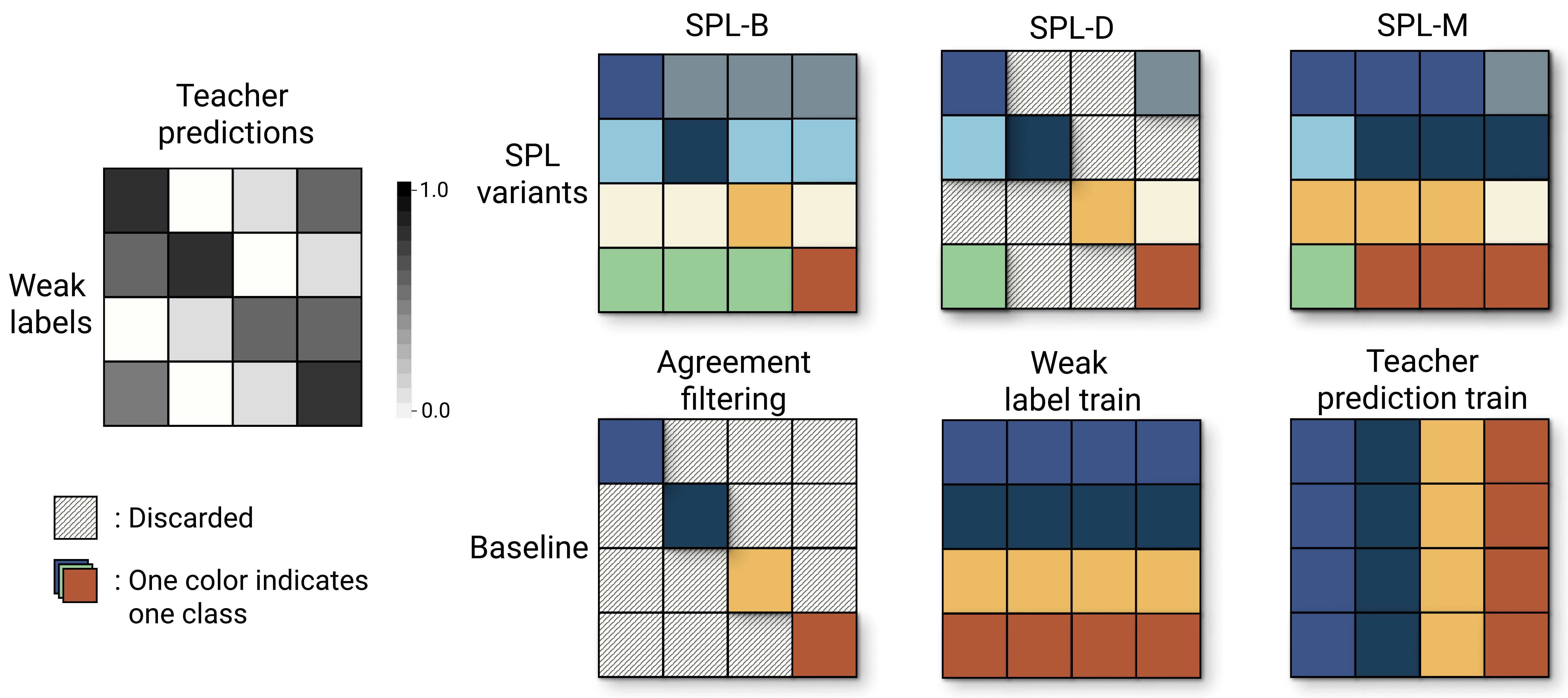} 
\caption{Illustration of the SPL mapping function. The left side is the confusion matrix between original weak labels and teacher predicted labels. The top-right illustrates three space-reduced variants of SPL. SPL-B: Using agreed and disagreed entries of each row as SPL classes. SPL-D: Using the top frequent entries as SPL classes. SPL-M: Merging less frequent off-diagonal entries to diagonals as SPL classes. The bottom-right illustrates three baseline pseudo-label strategies.} %See text for detailed explanations.
\label{fig:spl_assign_vis} 
\vspace{-.3cm}
\end{figure}

\textbf{Discard Tail Off-diagonal (SPL-D)}: The confusion matrix between weak labels and teacher predictions itself encodes information about label noises: the less frequent SPL classes have higher potentials to contain mislabeled data. 
Therefore, unlike SPL-M merging the un-selected 
%most infrequent \(N^2-K\)
classes to corresponding diagonals, SPL-D discards these training samples, resulting in a smaller yet potentially less noisy training dataset.

\textbf{Binary Merge (SPL-B)}: We explore using agreement and disagreement between weak labels and teacher predictions of video clips as a criterion to reduce the SPL space. In this case, the confusion matrix entries are reduced to \(2 \times N\) classes, including \(N\) in-diagonal classes (agreement) and \(N\) off-diagonal classes (disagreement) of each row. %This avoids creating too many fine-grained SPL classes that do not have sufficient samples. %The merged off-diagonal classes represent general concepts of related context that can be confused with other in-diagonal classes.

% The intuition behind is instead of creating too many fine-grained SPL classes, we can create one SPL for each original weak label. This SPL includes all related context that can not be assigned as the original weak label yet semantically related to it. 

% For the best of our knowledge, exploring the relation between weak labels and teacher predictions are not explored for weakly-supervised video recognition.

All variants can be viewed as pruning the confusion matrix, as illustrated in Figure \ref{fig:spl_assign_vis}. 
Why does the proposed simple pseudo labeling work? Figure \ref{fig:spl_assign_vis} provides a generalized view and intuitively unifies existing strategies as discussed in Section \ref{sc:method_overview}. Specifically, \emph{Weak label train} is the weakly-supervised training studied by \cite{ghadiyaram2019large}. \emph{Teacher prediction train} is a basic teacher-student self-training methods studied by \cite{furlanello2018born,xie2019self}. \emph{Agreement filtering} only takes samples whose the weak label is matched with the teacher model prediction on it. That being said, existing strategies more or less explore partial knowledge inside the confusion matrix. However, SPL considers higher-level class affinities that haven't been considered yet.
We will investigate these alternative strategies in experiments.

\section{Experiments}

% \subsection{Evaluation Datasets}\label{sc:dataset_setting} % and Evaluation Matrix

% The reason of chosing Kinetics-200 results in an optimal choice for future algorithm explorations by researchers with affordable resources along this critical direction. Moreover, this dataset has been treated as the main benchmark in recent NeurIPS papers e.g. [38], SDN (Choi, et al. NeurIPS'19),

We evaluate the proposed SPL algorithm on both common action recognition as well as fine-grained action recognition datasets. For the common action dataset, we mainly use Kinetics-200 (K200) \cite{xie2018rethinking} which is a subset of Kinetics-400 \cite{carreira2017quo}. In total, it contains 200 action categories with around 77K videos for training and 5K videos for validation.
Studies \cite{xie2018rethinking,choi2019can} show that evaluations on K200 can be well generalized to the full Kinetics and other cases.
Due to the huge computation resources required for extreme large-scale pre-training when taking full Kinetics as the target dataset, K200 results in an optimal choice for new algorithm explorations with an appropriate scale. Evaluation is conducted using Top-1 and Top-5 accuracy.
To validate the learned representations by SPL are generic and useful, we also follow recent works~\cite{miech2020end,patrick2020multi,stroud2020learning} to conduct experiments on popular HMDB-51~\cite{kuehne2011hmdb} and UCF-101~\cite{soomro2012ucf101} datasets. Specifically, we follow standard protocol to directly fine-tune the SPL pre-trained model on these two benchmarks to obtain results.

We also conduct complete evaluations on fine-grained action dataset SoccerNet \cite{giancola2018soccernet}, which is proposed for action event recognition in soccer broadcast videos and belongs to an important application domain.
It includes three foreground action classes: \textit{Goal, Yellow/Red Card, Substitution} and one \textit{Background} class for the rest contexts in soccer games.
We use 5547 video clips for training and 5547 clips for validation obtained from different full-match videos.
%we use 5547 video clips from XX long full-match videos for training and 5547 clips for validation  from XX long full-match videos.
For the evaluation matrix, we focus more on the performance of classifying foreground action classes, which are sparsely occurred in the broadcast videos. Therefore, mean average precision without background class %(denoted as \(\textrm{mAP\_no\_tn}\)) 
is adopted. % as the main evaluation matrix. 
% We also discuss cases for Something-Something \cite{goyal2017something} dataset in Appendix.
% We also discuss cases for dataset that has special class names in Section~\ref{sc:dataset_challenge} of Appendix.
We also discuss cases for dataset whose class names cannot be used as reliable search queries in Appendix. %(add section)

% in addition to the mean Average Precision (mAP), we focus more one the performance for the foreground action classes. Since to the characteristics of soccer broadcast videos that foreground actions are sparsely occurred,          

\subsection{Weakly-labeled data collection}\label{sc:data_collection}

To construct the pre-training dataset \(D_p\) for each target dataset (K200 and SoccerNet), we collect untrimmed web videos retrieved by a text-based search engine similar to \cite{caba2015activitynet,chen2015webly} and construct several dataset versions for following studies. Also see Appendix for more details. %We describe more details of obtaining the weakly-labeled data for each target dataset Mini-Knietic and Soccernet as follows.  

% \textbf{Web Kinetics-200.}
\textbf{WebK200.}
We treat the class names of Kinetics-200 dataset as the searching queries and use 4 languages for each query, including English, French, Spanish and Portuguese. We construct two web videos sets with different sizes: WebK200-147K-V with 147K videos and WebK200-285K-V with 285K videos (including more low-ranked videos returned by the search engine). Duplicate checking has been done on these two sets and videos with addresses appearing in the validation set of Kinetics-200 are removed. 
We sample a number of video clips with the length of 10 seconds from the retrieved videos. The number of clip samples for each class is roughly balanced according to the practice in \cite{ghadiyaram2019large}. 
%and 25 frame per second (fps) sample rate We will discuss more details about explorations of effect of video clips numbers in Section \ref{sc:exploration_results}.

% WebK200-147K-V with 147920 web videos (high threshold) and WebK200-285K-V with 285954 web videos(low threshold)

% \textbf{Soccernet.}
\textbf{WebS4.}
For the three foreground classes in SoccerNet: Goal, Yellow/Red Card, Substitution, 
we obtain the searching queries based on related terms from Wikipedia such as ``free kick goal'', ``corner kick goal'' resulting in 9 kinds of queries in total for these 3 foreground classes. For the Background class, we use ``soccer full match'' as the query. For each searching query, we use 3 languages including English, French and Spanish. We sample video clips with the length of 10 seconds and keep the number of clips for each class roughly balanced. Two web video sets are obtained with different number of total clips: WebS4-73K-C and WebS4-401K-C, where 73K and 401K represent the number of video clips in these two sets. 
%We conduct pre-training on these two sets and do fine-tuning on SoccerNet to obtain results on it. 
%and 10 fps sample rate from the retrieved videos 

\subsection{Implementation details}

We use 3D ResNet-50 \cite{wang2018non} with self-gating \cite{xie2018rethinking} as the baseline model and more details are described in the Appendix. Following \cite{wang2018non}, the network is initialized with ResNet-50 pre-trained on ImageNet \cite{deng2009imagenet}. At training stage, we use the batch size of 6 and take 16 RGB frames with temporal stride 4 as the input. The spatial size of each frame is \(224\times224\) pixels obtained from the same randomly cropping operation as \cite{wang2018non}. For the pre-training on our WebK200 sets, we set warm up training for 10 epochs with starting learning rate as 0.04 and then use learning rate 0.4 with cosine decay for 150 epochs. For the fine-tuning, we follow \cite{ghadiyaram2019large} to initialize the model from the last epoch of the pre-training and conduct end-to-end fine-tuning. We set warm up training for 10 epochs with starting learning rate as 0.04 and then use learning rate of 0.4 with cosine decay for 60 epochs. For SoccerNet dataset, we use the same pre-training setting with WebK200 to conduct pre-training on the WebS4 sets. For the fine-tuning, we use learning rate of 0.005 for 20 epochs. More settings of hyper-parameters are described in the appendix. They are obtained to get the best performance of baselines. Our method is implemented using TensorFlow \cite{tensorflow2015}.

%Synchronous stochastic gradient descent (SGD) is used to train the model. 

\begin{table}
\centering
    \scalebox{1.0}{
    \begin{tabular}{l|c|c}
        \toprule
        %  \multicolumn{2}{c}{Part}                   \\
        % \cmidrule(r){1-2}
        Pre-train Method & Top-1  & Top-5\\
        \toprule
        ImageNet Pre-train & 80.6 & 94.7 \\
        Weak Label Train \cite{ghadiyaram2019large} &  82.8 & 95.6  \\
        Teacher Prediction Train \cite{xie2019self} & 81.9 & 95.0 \\
        Agreement Filtering Train & 82.9 & 95.4 \\ %\midrule
        Data Parameters~\cite{saxena2019data} & 83.2 & 95.5 \\
        % WLT &  82.8 & 95.6  \\
        % TPT & 81.9 & 95.0 \\
        % AFT & 82.9 & 95.4 \\ %\midrule
        % DP & 83.2 & 95.5 \\        
        \midrule
        SPL-B (Ours) & \textbf{84.3} & \textbf{95.7} \\
        \bottomrule
    \end{tabular}}
\vspace{-3mm}
\caption{Comparisons with different pre-training strategies on WebK200-147K-V set with \(6.7 \times 10^{5}\) clips. Fine-tuning results on Kinetics-200 dataset are shown.}
\label{tab:minik200_label_sapce}
\vspace{-3mm}
\end{table}

\begin{figure}
\centering
\includegraphics[width=0.85\linewidth]{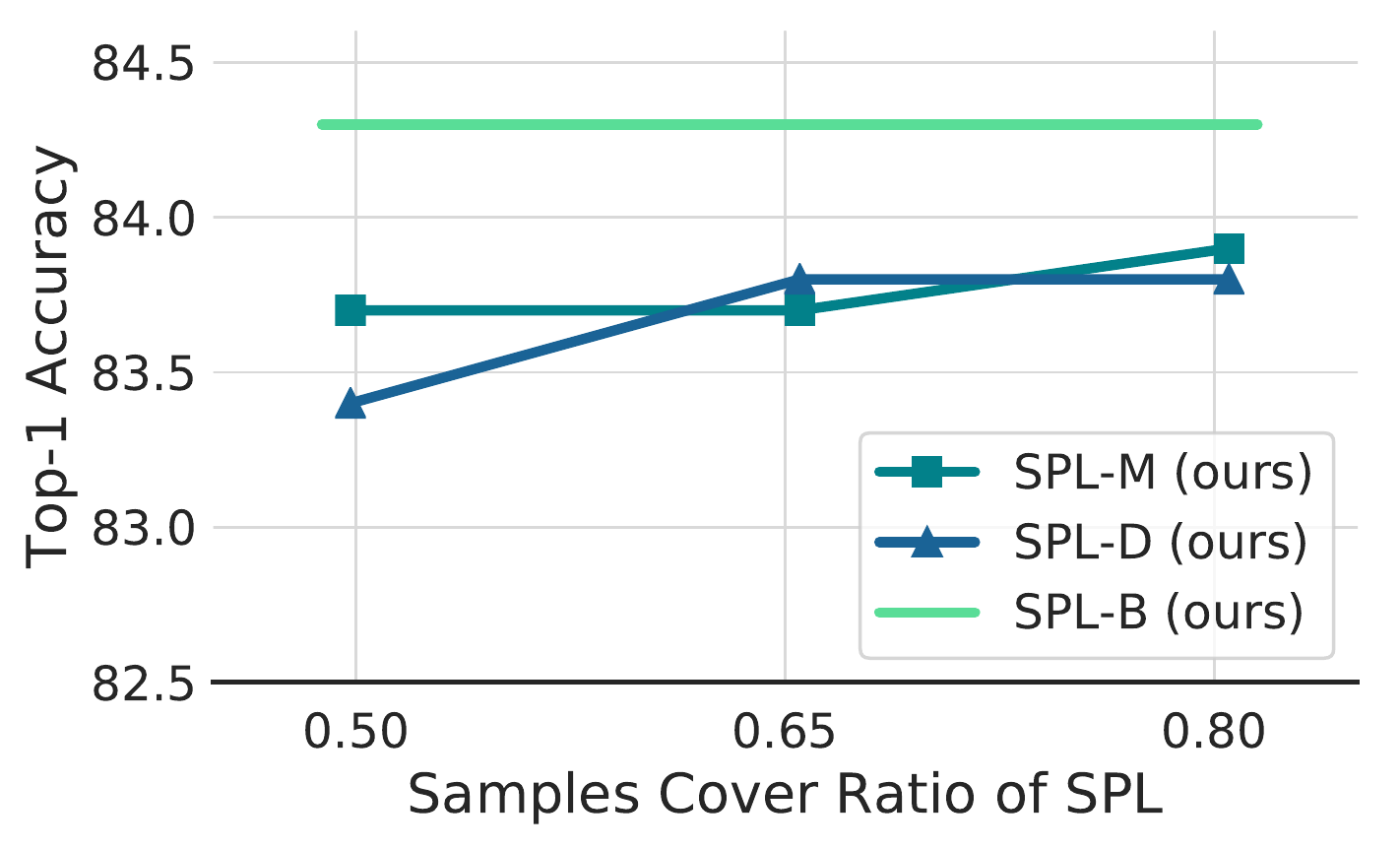} 
\vspace{-3mm}
\caption{SPL-M and SPL-D with different samples cover ratio (SCR) defined in the Section \ref{sc:results_miniK200} for pre-training on WebK200-147K-V with \(6.7 \times 10^{5}\) clips. Fine-tuning results on Kinetics-200 dataset are shown.} %SPL-M and SPL-D with different samples cover ratio (SCR) defined in the Section \ref{sc:results_miniK200}
\label{fig:result_spl_number}  
\vspace{-4mm}
\end{figure}

% \zizhao{Instead of trim content, it worthy to reduce the space of followings by either adding space-margin or remove subsubsection is necessary. Let's considering remove content at last day.}

\subsection{Results on the Kinetics-200 dataset}\label{sc:results_miniK200}

In this section, we verify the effectiveness of the proposed method on Kinetics-200 via studies of different perspectives and explorations. Fine-tuning results are reported. 

% \subsubsection{Comparison of Different Pre-training Strategies}
% Section \ref{sc:variations} and Figure \ref{fig:spl_assign_vis} categorize different pseudo labeling strategies, including existing baseline methods Weak Label and Teacher Prediction widely used by weakly-supervised pre-training as well as the variants of SPL. Here we compare these strategies to ours.  

% \textbf{Comparisons with baselines.} 
\textbf{Comparisons with other pre-training strategies.}
Section \ref{sc:variations} and Figure \ref{fig:spl_assign_vis} categorize different pseudo-label strategies. Here we compare these strategies to ours.  
We report results based on their pre-training on our WebK200-147K-V set with \(6.7 \times 10^{5}\) clips. From Table \ref{tab:minik200_label_sapce}, we find they can all improve upon the baseline ImageNet pre-training. The performance gap between pre-training using Weak Label \cite{ghadiyaram2019large} and Teacher Prediction \cite{xie2019self} indicates there are more useful information included in weak labels. 
Although Agreement Filtering can do some noise reduction to the web videos, it discards around 60\% of training samples resulting in a comparable performance with Weak Label. 
We also adopt Data Parameters \cite{saxena2019data}, one of the recent state-of-the-art methods for learning with noisy labels, to conduct pre-training on web videos. %Compared with them,   
Our SPL-B (variation with the best performance on Kinetics-200) outperforms these baselines and is able to take use of all noisy data.

% \emph{Weak label train} is the weakly-supervised training studied by \cite{ghadiyaram2019large}. \emph{Teacher prediction train} is a basic teacher-student distillation methods studied by \cite{furlanello2018born,xie2019self}.

\begin{figure}
\centering
\includegraphics[width=0.85\linewidth]{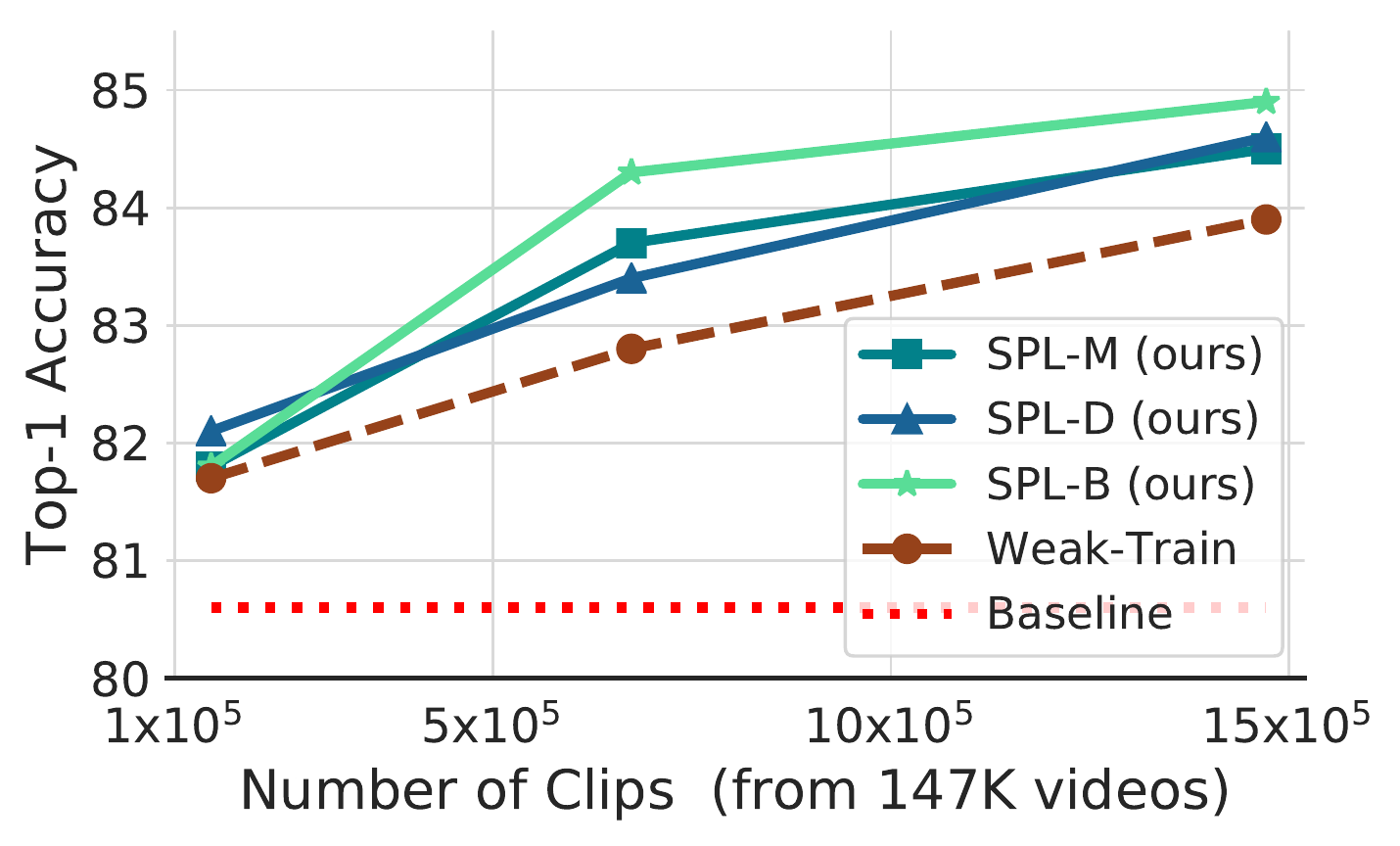} 
\vspace{-3mm}
\caption{Effect of different numbers of clips given a fixed number of videos (WebK200-147K-V). Fine-tuning results on Kinetics-200 dataset are shown.} %SPL-M and SPL-D with different samples cover ratio (SCR) defined in the Section \ref{sc:results_miniK200}
\label{fig:result_clip_number}  
\vspace{-3mm}
\end{figure}

\begin{table}
\centering
    \scalebox{0.95}{
    \begin{tabular}{l|c|c}
        \toprule
        \multirow{2}{*}{Pre-train Method} & \multicolumn{2}{c}{Number of videos} \\ \cmidrule{2-3}
        & 147K-V & 285K-V  \\
        \midrule
        Synthetic Noise Ratio \tablefootnote{We use \(\frac{\text{\# of off-diagonal elements}}{\text{\# of total elements}}\) in the confusion matrix as a synthetic measurement of the noise ratio.}%Given no ground-truth, 
        & 58.9 \% &  65.5\% \\
        \midrule
        Weak Label Train \cite{ghadiyaram2019large} & 83.9 & 84.0   \\
        SPL-M (Ours) & 84.5  & 84.8  \\
        SPL-D (Ours) & 84.6  & 84.9  \\
        SPL-B (Ours) & \textbf{84.9}  &  \textbf{85.3}  \\
        \bottomrule
    \end{tabular}
    }
\vspace{-2mm}
\caption{Effect of different number of videos given a fixed number of clips (around \(1.4 \times 10^{6}\)). Results are reported on Kinetics-200 (Top-1 accuracy).}
\label{tab:minik200_video_numbers}
\vspace{-3mm}
\end{table}

% \textbf{Effect of the number of SPL classes.}
\textbf{Comparisons between different variations of SPL.}
% Our SPL-M and SPL-D can be set with different numbers of SPL classes as a reduction of the \(O(N^2)\) label space which is 40000 for the Mini-Kinetics dataset. 
To compare the performance of the variants of SPL, 
%including SPL-M, SPL-D and SPL-B, 
we conduct experiments on our WebK200-147K-V set with \(6.7 \times 10^{5}\) clips for pre-training. We start with total number of SPL classes as \(K \times N=400\) so that the label space is consistent for the three variations. 
The label space of SPL-D and SPL-B is controlled by hyper-parameter \(K\) and their space is reduced by merging or discarding samples belong to infrequent SPLs. There is a question about how many frequent SPLs to keep. More classes introduce more fine-grained tail SPLs yet higher computation cost. We define samples cover ratio (SCR) = \(\frac{\text{\# of samples in selected SPLs}}{\text{\# of total samples}}\).
Specifically, 400 SPL classes give SCR\(=49.81\%\). We evenly increase SCR by 15\% to get 1600 and 4500 SPL classes with SCR of 65\% and 80\% respectively. From the result in Figure \ref{fig:result_spl_number}, we find that including more SPL classes can generally improve the performance of SPL-M and SPL-D. But the overall improvement gain is limited and they underperform SPL-M. 

% \subsubsection{Effect of the number of training samples, more clips or more videos?}
\textbf{Effect of the number of training samples, more clips or more videos?}
It is a common practice to improve the performance of deep neural networks by increasing the number of training samples \cite{mahajan2018exploring,ghadiyaram2019large}. 
There are two practical options to do so given untrimmed web videos: (1) sampling more clips from a given number of web videos or (2) collecting more videos to sample clips. Both ways have potential drawbacks. The first one may result in duplication of visual contents. The second one may lead to lower quality of weak labels because this means we have to include more low-ranked videos returned by the search engine. %where correlations between the text query and the video content are decreasing.  
In the following experiments, 
% we are not using datasets as large as \cite{ghadiyaram2019large,yalniz2019billion} use (up to 100x web videos), as the goal here is not to demonstrate extreme accuracy via massive data. Instead, 
we aim to study this practical question and verify the effectiveness of SPL.

\begin{table}
  \centering
  \scalebox{0.9}{
  \begin{tabular}{l|c|c}
    \toprule
    Method      & Top-1  & Top-5\\
    \midrule
    I3D$_{(\rm{CVPR'17})}$~\cite{carreira2017quo} &   78.0  & - \\ 
    S3D$_{(\rm{ECCV'18})}$~\cite{xie2018rethinking}    & 78.4  & - \\ 
    R3D-50$_{(\rm{CVPR'18})}$~\cite{wang2018non}  & 75.5  & 92.2 \\
    R3D-50-NL$_{(\rm{CVPR'18})}$~\cite{wang2018non}    & 77.5  & 94.0 \\
    R3D-50-CGNL$_{(\rm{NeurIPS'18})}$~\cite{yue2018compact}   & 78.8 & 94.4 \\
    TBN$_{(\rm{AAAI'19})}$~\cite{li2019temporal} & 69.5 & 89.4 \\
    \midrule
    SPL (Ours)   & \textbf{85.3} & \textbf{96.6} \\
    \bottomrule
  \end{tabular}
  }
   \vspace{-2mm}
  \caption{Results of other methods on Kinetics-200.}
  \label{tab:minik200_sota}
  \vspace{-5mm}
\end{table}

\begin{table*}
  \centering
  \scalebox{1.0}{
  \begin{tabular}{l|c|c|c|c|c|c}
    \toprule
    % Method     & Baseline & WLT & AF & TPT & SPL (Ours)   \\
    Method     & Extra data & Video \#  & Model & Modality & HMDB-51 & UCF-101   \\
    \midrule
    S3D-G$_{(\rm{ECCV'18})}$~\cite{xie2018rethinking}  & ImageNet & - &S3D-G & V & 57.7 & 86.6  \\
    R(2+1)D$_{(\rm{ECCV'18})}$~\cite{tran2018closer} & ImageNet & - & R(2+1)D & V & 48.1 & 84.0  \\
    \midrule
    OPN$_{(\rm{ICCV'17})}$~\cite{lee2017unsupervised}  & UCF & 13K & VGG & V & 23.8 & 59.6 \\ 
    ST-Puzzle$_{(\rm{AAAI'19})}$~\cite{kim2019self}  & K400 & 240K & R3D-18 & V & 33.7 & 63.9\\ 
   SpeedNet$_{(\rm{CVPR'20})}$~\cite{benaim2020speednet}  & K400 & 240K  & S3D-G & V & 48.8 & 81.1\\
   MemDPC$_{(\rm{ECCV'20})}$~\cite{han2020memory}  & K400 & 240K & R-2D3D & V & 54.5 & 86.1 \\
    CPD$_{(\rm{arXiv'20})}$~\cite{li2020learning}  & K400 & 240K & R3D-50 & V & 58.4 & 88.7\\ 
    AVTS$_{(\rm{NeurIPS'18})}$~\cite{korbar2018cooperative} & Audioset & 2M & MC3 & V+A & 61.6 & 89.0\\
    XDC$_{(\rm{arXiv'19})}$~\cite{alwassel2019self}  & K400 & 240K & R(2+1)D & V+A & 47.1 & 84.2\\
    XDC$_{(\rm{arXiv'19})}$~\cite{alwassel2019self}  & IG65M & 65M & R(2+1)D & V+A & 67.4 & 94.2\\
    GDT$_{(\rm{arXiv'20})}$~\cite{patrick2020multi}  & K400 & 240K & R(2+1)D & V+A & 57.8 & 88.7 \\
    CBT$_{(\rm{arXiv'19})}$~\cite{sun2019learning}  & K600 & 390K & S3D & V+T & 44.5 & 79.5\\ 
   MIL-NCE$_{(\rm{CVPR'20})}$~\cite{miech2020end}  & HT100M & 1.2M & S3D & V+T & 61.0 & 91.3\\
   WVT$_{(\rm{arXiv'20})}$~\cite{stroud2020learning}  & WVT-70M & 70M & S3D-G & V+T & 65.3 & 90.3\\
   \midrule
   Our baseline & ImageNet & - & R3D-50-G & V & 46.0 & 84.9\\
   SPL (Ours) & WebK200-147K & 147K & R3D-50-G & V& \textbf{67.6}& \textbf{94.6} \\
   \bottomrule 
  \end{tabular} }
   \vspace{-3mm}
  \caption{More complete comparisons on HMDB-51 and UCF-101 benchmarks. For modality, ``V'' refers to visual only, ``A'' represents audio and ``T'' means text description.} 
   \label{tab:ucf_hmdb_complete}
   \vspace{-5mm}
\end{table*} % 

\emph{Effect of more clips.}
%1 clip from each video or uniformly sampling 5 or 10 clips from each video in
We sample different numbers of clips from WebK200-147K-V set (described in Section \ref{sc:data_collection}) and plot results of different pre-training strategies in Figure \ref{fig:result_clip_number}. 
%The total number of SPL classes is set to 400 for all three versions. 
The baseline result with red dot line represents the performance of using ImageNet pre-trained models. Results show that sampling more video clips from a given number of untrimmed videos can help improve the model performance. 
We also find that with a sufficient number of video clips available, our SPL methods consistently outperform weak label pre-training by providing rich and valid supervision knowledge.

\emph{Effect of more videos.} %As described in Section \ref{sc:data_collection}, we have collected two web videos sets: WebK200-147K-V with 147K web videos and WebK200-285K-V with 285K web videos. 
We sample a similar number of total video clips, around \(1.4 \times 10^6\), from WebK200-147K-V (147K videos) and WebK200-285K-V (285K videos) to obtain two training sets. We conduct teacher model inference on these two sets to get a synthetic measurement of the noise ratio and find this ratio is larger in WebK200-285K-V set. %This confirms our assumption above that the quality of weak labels decreases with increasing of videos. 
The comparison in Table \ref{tab:minik200_video_numbers} indicates that, 
% for all compared methods, using more videos can bring higher performance gain. This indicates 
though synthetic noise ratio is higher with the increase of videos, enriched visual contents are beneficial to some extent.
%, we find that, in general, increasing number of videos can bring higher performance gain. %This observation matches the conclusion of \cite{ghadiyaram2019large}. 
More importantly, our SPL-B obtains more performance gain 
%(0.4\% v.s. 0.1\% based on an already high performance on 147K set) 
than directly using weak labels, which also indicates its robustness to label noise.
%For different variations of SPL when pre-trained on WebK200-285K-V, SPL-B achieves the best performance with 85.3\% top-1 accuracy on Kinetics-200 validation set, which is 1.3\% higher than the weak label pre-training and 4.7\% higher the ImageNet pre-training. These improvements are significant on Kinetics-200 dataset according to the practice in \cite{yue2018compact}. 

\textbf{Comparisons with other methods on Kinetics-200.}
In Table \ref{tab:minik200_sota}, we list results of other existing methods on this benchmark: I3D~\cite{carreira2017quo}, S3D~\cite{xie2018rethinking}, R3D-NL~\cite{wang2018non}, R3D-CGNL~\cite{yue2018compact} and TBN~\cite{li2019temporal}. The results show that our method, which uses only extra 2-4x more web videos, is able to outperform the previous best reported number by a large margin (over 5\%).    

\begin{table}
\centering
    \scalebox{0.9}{
    \begin{tabular}{l|c|c}
        \toprule
        \multirow{2}{*}{Pre-train Method} & \multicolumn{2}{c}{Pre-train Set} \\ \cmidrule{2-3}
        & WebS4-73K-C & WebS4-401K-C   \\
        \midrule
        Baseline & 73.7 & 73.7  \\
        Weak Label Train~\cite{ghadiyaram2019large} &  74.8  &  75.3 \\
        Agreement Filter & 75.1 & 75.4  \\
        Teacher Predication~\cite{xie2019self} & 74.1 &  75.2 \\
         \midrule
         SPL (Ours)  & \textbf{76.1} &  \textbf{76.8}\\
        \bottomrule
    \end{tabular}
    }
 \vspace{-3mm}
\caption{Results on the SoccerNet dataset. ``Baseline'' represents the ImageNet pre-training.}
\label{tab:soccernet}
\vspace{-4mm}
\end{table}

\begin{table}
\centering
  \scalebox{1.0}{
  \begin{tabular}{l|c}
    \toprule
    % \multicolumn{2}{c}{Part}                   \\
    % \cmidrule(r){1-2}
    Method   & Top-1 ($\%$)  \\
    \midrule
    None$_{(\rm{CVPR'17})}$~\cite{patrini2017making}  & 79.43 \\ 
    Forward$_{(\rm{CVPR'17})}$~\cite{patrini2017making}  & 80.38 \\ 
    CleanNet$_{(\rm{CVPR'18})}$~\cite{lee2018cleannet}   &  79.90 \\
    NoiseRank$_{(\rm{ECCV'20})}$~\cite{sharma2020noiserank}  & 79.57 \\
    \midrule
    SPL (Ours)  & \textbf{80.50}\\
   \bottomrule
    \hline
  \end{tabular}
  }
\vspace{-3mm}
\caption{Results on the Clothing1M~\cite{xiao2015learning} image classification datasets. All methods use the ResNet-50 backbone.}
\label{tab:clothing1m}
\vspace{-6mm}
\end{table}

\begin{figure*}
\centering
\includegraphics[width=0.85\linewidth]{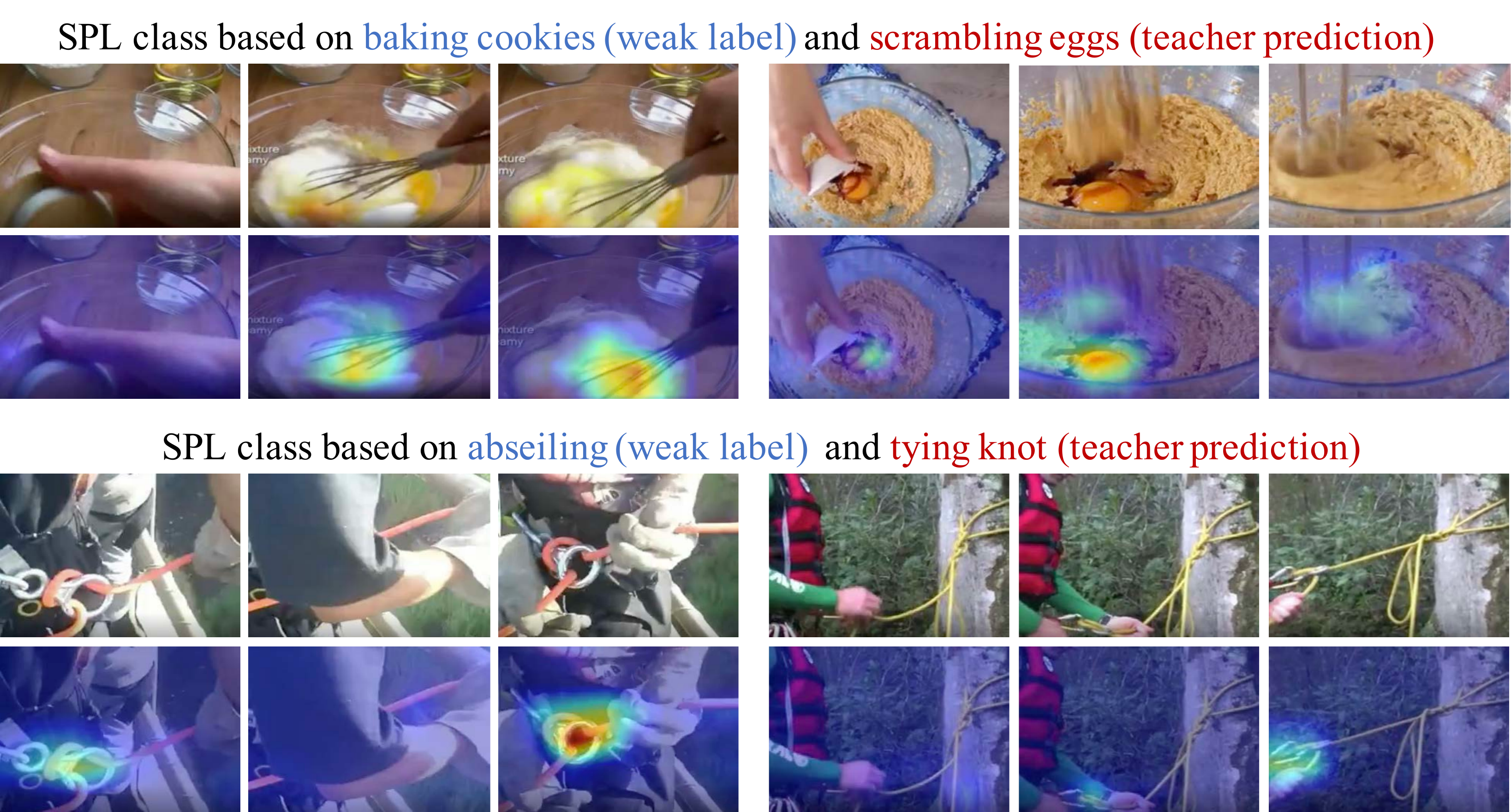} 
\vspace{-2mm}
\caption{Examples of attention visualization for SPL classes. Original weak label (blue) and the teacher model prediction (red) are listed. Some meaningful ``middle ground'' concepts can be learnt by SPL, such as mixing up the eggs and flour (top) and using the abseiling equipment (bottom). }
\label{fig:spl_example} 
\vspace{-5mm}
\end{figure*}

\textbf{Attention visualization of SPL classes.}
We visualize the visual concepts learned from SPLs via attention visualization. Specifically, we extend Grad-CAM \cite{selvaraju2017grad} to 3D CNNs to show the model's focus when making predictions. 
In Figure \ref{fig:spl_example}, we show some examples of SPL classes along with attention maps of the model trained using SPL. 
%Original weak label (in blue) and the teacher model prediction (in red) are also listed. 
It is interesting to observe some meaningful ``middle ground'' concepts can be learnt by SPL, such as mixing the eggs and flour, using the abseiling equipment.

\subsection{Evaluation of learned representation by fine-tuning HMDB-51 and UCF-101}\label{sc:results_hmdbucf}

To validate the learned representations by SPL are generic and useful, we further follow recent works~\cite{miech2020end,patrick2020multi,stroud2020learning} to conduct experiments on HMDB51 and UCF101 datasets. Note that we directly fine-tune the SPL-B pre-trained model from Webk200-285K-V set on these two benchmarks to obtain results - during this process, no new noisy web dataset is collected when treating HMDB-51 and UCF-101 datasets as the target dataset, which is a common way to evaluate the learned representations for large-scale web video training \cite{stroud2020learning}. In Table \ref{tab:ucf_hmdb_complete}, we list comparisons with recent self-supervised and webly-supervised pre-training methods that relies on different modalities: OPN~\cite{lee2017unsupervised}, ST-Puzzle~\cite{kim2019self}, CBT~\cite{sun2019learning}, SpeedNet~\cite{benaim2020speednet}, MemDPC~\cite{han2020memory}, CPD~\cite{li2020learning}, AVTS~\cite{korbar2018cooperative}, XDC~\cite{alwassel2019self}, GDT~\cite{patrick2020multi}, MIL-NCE~\cite{miech2020end}, WVT~\cite{stroud2020learning}. We also report results of our ImageNet initialized baselines. 

It is worth to note that it is usually hard to conduct totally direct comparisons between recent webly-supervised video understanding frameworks. This is duo to the public unavailable web data used in each work, e.g. IG65M~\cite{alwassel2019self,ghadiyaram2019large}, WVT-70M~\cite{stroud2020learning} and huge cost for compution resources. In such case, we find SPL achieves competitive results by using a smaller number of videos to conduct the pre-training. This is also achieved by improving upon a weaker baseline model compared with R(2+1)D and S3D-G initialized by ImageNet pre-train weights.

 % we find SPL improves our baseline significantly and also outperforms recent self and webly-supervised pre-training methods that relies on only visual modality: MemDPC~\cite{han2020memory}, SpeedNet~\cite{benaim2020speednet}, CPD~\cite{li2020learning}, MIL-NCE~\cite{miech2020end}, WVT~\cite{stroud2020learning}. Our method also uses a smaller number of videos compared with them. %More complete comparisons are in Table \ref{tab:ucf_hmdb_complete} of Appendix, where we include results of more methods, settings of them and more discussions. % More complete comparision are in Table of appendix， modality, number of video in pre-train set, etc. 

% \vspace{-1mm}
\subsection{Experiments on SoccerNet dataset}
We also conduct experiments on SoccerNet \cite{giancola2018soccernet}, a fine-grained action recognition dataset. 
Different from Kinetics-200 actions, this dataset contains broadcast videos from soccer matches, so all classes contain sports actions sharing very similar visual (background) content. Therefore there exists high confusion between different classes. Moreover, we find actions in untrimmed web videos is transitory, leading to high temporal noise. 
We use two web video sets WebS4-73K-C and WebS4-401K-C with 73K clips and 401K clips respectively as described in Section \ref{sc:data_collection}. Since the label space is not high, we use the full version of SPL that generates $4^2$ SPL classes. In Table \ref{tab:soccernet}, we show the fine-tuning results on SoccerNet val set based on different types of pre-training. %``Baseline'' represents the ImageNet pre-training. 
Our SPL method consistently outperforms other pre-training strategies, suggesting the advantage of SPL to learn short actions from  untrimmed web videos.

% \subsection{Generalizing SPL to Weakly-labeled Image Data}
\subsection{Generalizing SPL to image classification}
We study whether our proposed SPL has generalization ability to other domains other than videos. 
We test it on Clothing1M \cite{xiao2015learning}, a large-scale image datasets with real-world label noises. Clothing1M contains $47,570$ training images with human-annotated labels and $\sim$1M images with noisy labels. There are 14 original fashion classes. This dataset is challenging and $1\%$ improvement is regarded as important. Different methods are studied to utilize human-annotated to better train on large weakly-labeled web data. SPL can directly be used for this task.
Since the label space is not high, we use the basic version of SPL that generates $14^2$ SPLs for pre-training on the $\sim$1M images. 
We follow common experimental setting \cite{lee2018cleannet} and starts ResNet-50 pre-training with random initialization. Then we fine-tune on the clean set. Table \ref{tab:clothing1m} compares against previous methods, suggesting generalizability of the proposed method
% such as None~\cite{patrini2017making}, Forward~\cite{patrini2017making}, CleanNet~\cite{lee2018cleannet}, NoiseRank~\cite{sharma2020noiserank}. 
(source code will be provided\footnote{We understand reproduce large-scale video training is costly. To encourage reproducibility of our method, we will provide the source code of image recognition experiments.}). 
SPL either outperforms or achieves competitive results.

\section{Conclusion}

We propose a novel and particularly simple method of exploring SPLs from untrimmed weakly-labeled web videos. Although previous literature has shown large-scale pre-training with weak labels can directly benefit, we demonstrate that SPLs can provide enriched supervision and bring much larger improvements. 
Importantly, SPL does not increase training complexity and can be treated as an off-the-shelf technique to integrate with teacher-student like training frameworks in orthogonal. 
In addition, we believe it is promising direction to discover meaningful visual concepts by bridging weak labels and the knowledge distilled from teacher models. 
SPL also demonstrates promising generalization to the image recognition domain, suggesting promising future directions like applying SPL to other tasks where there exists uncertainty in labels. 

\section{Acknowledgement}
We thank Debidatta Dwibedi, David A Ross, Chen Sun, Jonathan C. Stroud for their valuable comments and help on this work.

{\small
\balance
\bibliographystyle{ieee_fullname}
\bibliography{egbib}

\begin{thebibliography}{10}\itemsep=-1pt

\bibitem{tensorflow2015}
Mart\'{\i}n Abadi, Ashish Agarwal, Paul Barham, Eugene Brevdo, Zhifeng Chen,
  Craig Citro, Greg~S. Corrado, Andy Davis, Jeffrey Dean, Matthieu Devin,
  Sanjay Ghemawat, Ian Goodfellow, Andrew Harp, Geoffrey Irving, Michael Isard,
  Yangqing Jia, Rafal Jozefowicz, Lukasz Kaiser, Manjunath Kudlur, Josh
  Levenberg, Dan Man\'{e}, Rajat Monga, Sherry Moore, Derek Murray, Chris Olah,
  Mike Schuster, Jonathon Shlens, Benoit Steiner, Ilya Sutskever, Kunal Talwar,
  Paul Tucker, Vincent Vanhoucke, Vijay Vasudevan, Fernanda Vi\'{e}gas, Oriol
  Vinyals, Pete Warden, Martin Wattenberg, Martin Wicke, Yuan Yu, and Xiaoqiang
  Zheng.
\newblock {TensorFlow}: Large-scale machine learning on heterogeneous systems,
  2015.
\newblock Software available from tensorflow.org.

\bibitem{alwassel2019self}
Humam Alwassel, Dhruv Mahajan, Lorenzo Torresani, Bernard Ghanem, and Du Tran.
\newblock Self-supervised learning by cross-modal audio-video clustering.
\newblock {\em arXiv:1911.12667}, 2019.

\bibitem{ba2014deep}
Jimmy Ba and Rich Caruana.
\newblock Do deep nets really need to be deep?
\newblock In {\em NeurIPS}, 2014.

\bibitem{benaim2020speednet}
Sagie Benaim, Ariel Ephrat, Oran Lang, Inbar Mosseri, William~T Freeman,
  Michael Rubinstein, Michal Irani, and Tali Dekel.
\newblock Speednet: Learning the speediness in videos.
\newblock In {\em CVPR}, 2020.

\bibitem{bucilu2006model}
Cristian Buciluǎ, Rich Caruana, and Alexandru Niculescu-Mizil.
\newblock Model compression.
\newblock In {\em KDD}, 2006.

\bibitem{caba2015activitynet}
Fabian Caba~Heilbron, Victor Escorcia, Bernard Ghanem, and Juan Carlos~Niebles.
\newblock Activitynet: A large-scale video benchmark for human activity
  understanding.
\newblock In {\em CVPR}, 2015.

\bibitem{carreira2017quo}
Joao Carreira and Andrew Zisserman.
\newblock Quo vadis, action recognition? a new model and the kinetics dataset.
\newblock In {\em CVPR}, 2017.

\bibitem{chang2019d3tw}
Chien-Yi Chang, De-An Huang, Yanan Sui, Li Fei-Fei, and Juan~Carlos Niebles.
\newblock D3tw: Discriminative differentiable dynamic time warping for weakly
  supervised action alignment and segmentation.
\newblock In {\em CVPR}, 2019.

\bibitem{chen2020simple}
Ting Chen, Simon Kornblith, Mohammad Norouzi, and Geoffrey Hinton.
\newblock A simple framework for contrastive learning of visual
  representations.
\newblock {\em arXiv:2002.05709}, 2020.

\bibitem{chen2015webly}
Xinlei Chen and Abhinav Gupta.
\newblock Webly supervised learning of convolutional networks.
\newblock In {\em ICCV}, 2015.

\bibitem{chen2013neil}
Xinlei Chen, Abhinav Shrivastava, and Abhinav Gupta.
\newblock Neil: Extracting visual knowledge from web data.
\newblock In {\em ICCV}, 2013.

\bibitem{choi2019can}
Jinwoo Choi, Chen Gao, Joseph~CE Messou, and Jia-Bin Huang.
\newblock Why can't i dance in the mall? learning to mitigate scene bias in
  action recognition.
\newblock In {\em NeurIPS}, 2019.

\bibitem{deng2009imagenet}
Jia Deng, Wei Dong, Richard Socher, Li-Jia Li, Kai Li, and Li Fei-Fei.
\newblock Imagenet: A large-scale hierarchical image database.
\newblock In {\em CVPR}, 2009.

\bibitem{divvala2014learning}
Santosh~K Divvala, Ali Farhadi, and Carlos Guestrin.
\newblock Learning everything about anything: Webly-supervised visual concept
  learning.
\newblock In {\em CVPR}, 2014.

\bibitem{duan2020omni}
Haodong Duan, Yue Zhao, Yuanjun Xiong, Wentao Liu, and Dahua Lin.
\newblock Omni-sourced webly-supervised learning for video recognition.
\newblock {\em arXiv:2003.13042}, 2020.

\bibitem{duan2012exploiting}
Lixin Duan, Dong Xu, and Shih-Fu Chang.
\newblock Exploiting web images for event recognition in consumer videos: A
  multiple source domain adaptation approach.
\newblock In {\em CVPR}, 2012.

\bibitem{feichtenhofer2019slowfast}
Christoph Feichtenhofer, Haoqi Fan, Jitendra Malik, and Kaiming He.
\newblock Slowfast networks for video recognition.
\newblock In {\em ICCV}, 2019.

\bibitem{feichtenhofer2016convolutional}
Christoph Feichtenhofer, Axel Pinz, and Andrew Zisserman.
\newblock Convolutional two-stream network fusion for video action recognition.
\newblock In {\em CVPR}, 2016.

\bibitem{furlanello2018born}
Tommaso Furlanello, Zachary~C Lipton, Michael Tschannen, Laurent Itti, and
  Anima Anandkumar.
\newblock Born again neural networks.
\newblock In {\em ICML}, 2018.

\bibitem{gan2016webly}
Chuang Gan, Chen Sun, Lixin Duan, and Boqing Gong.
\newblock Webly-supervised video recognition by mutually voting for relevant
  web images and web video frames.
\newblock In {\em ECCV}, 2016.

\bibitem{gan2016you}
Chuang Gan, Ting Yao, Kuiyuan Yang, Yi Yang, and Tao Mei.
\newblock You lead, we exceed: Labor-free video concept learning by jointly
  exploiting web videos and images.
\newblock In {\em CVPR}, 2016.

\bibitem{ghadiyaram2019large}
Deepti Ghadiyaram, Du Tran, and Dhruv Mahajan.
\newblock Large-scale weakly-supervised pre-training for video action
  recognition.
\newblock In {\em CVPR}, 2019.

\bibitem{giancola2018soccernet}
Silvio Giancola, Mohieddine Amine, Tarek Dghaily, and Bernard Ghanem.
\newblock Soccernet: A scalable dataset for action spotting in soccer videos.
\newblock In {\em CVPR Workshops}, 2018.

\bibitem{goyal2017something}
Raghav Goyal, Samira~Ebrahimi Kahou, Vincent Michalski, Joanna Materzynska,
  Susanne Westphal, Heuna Kim, Valentin Haenel, Ingo Fruend, Peter Yianilos,
  Moritz Mueller-Freitag, et~al.
\newblock The" something something" video database for learning and evaluating
  visual common sense.
\newblock In {\em ICCV}, 2017.

\bibitem{han2020memory}
Tengda Han, Weidi Xie, and Andrew Zisserman.
\newblock Memory-augmented dense predictive coding for video representation
  learning.
\newblock {\em ECCV}, 2020.

\bibitem{he2016deep}
Kaiming He, Xiangyu Zhang, and Jian Sun.
\newblock Deep residual learning for image recognition.
\newblock In {\em CVPR}, 2016.

\bibitem{hinton2015distilling}
Geoffrey Hinton, Oriol Vinyals, and Jeff Dean.
\newblock Distilling the knowledge in a neural network.
\newblock In {\em NeurIPS Workshop}, 2015.

\bibitem{hinton2012improving}
Geoffrey~E Hinton, Nitish Srivastava, Alex Krizhevsky, Ilya Sutskever, and
  Ruslan~R Salakhutdinov.
\newblock Improving neural networks by preventing co-adaptation of feature
  detectors.
\newblock {\em arXiv:1207.0580}, 2012.

\bibitem{ioffe2015batch}
Sergey Ioffe and Christian Szegedy.
\newblock Batch normalization: Accelerating deep network training by reducing
  internal covariate shift.
\newblock In {\em ICML}, 2015.

\bibitem{kim2019self}
Dahun Kim, Donghyeon Cho, and In~So Kweon.
\newblock Self-supervised video representation learning with space-time cubic
  puzzles.
\newblock In {\em AAAI}, 2019.

\bibitem{korbar2018cooperative}
Bruno Korbar, Du Tran, and Lorenzo Torresani.
\newblock Cooperative learning of audio and video models from self-supervised
  synchronization.
\newblock In {\em NeurIPS}, 2018.

\bibitem{krizhevsky2012imagenet}
Alex Krizhevsky, Ilya Sutskever, and Geoffrey~E Hinton.
\newblock Imagenet classification with deep convolutional neural networks.
\newblock In {\em NeurIPS}, 2012.

\bibitem{kuehne2019mining}
Hilde Kuehne, Ahsan Iqbal, Alexander Richard, and Juergen Gall.
\newblock Mining youtube-a dataset for learning fine-grained action concepts
  from webly supervised video data.
\newblock {\em arXiv:1906.01012}, 2019.

\bibitem{kuehne2011hmdb}
Hildegard Kuehne, Hueihan Jhuang, Est{\'\i}baliz Garrote, Tomaso Poggio, and
  Thomas Serre.
\newblock Hmdb: a large video database for human motion recognition.
\newblock In {\em ICCV}, 2011.

\bibitem{laptev2005space}
Ivan Laptev.
\newblock On space-time interest points.
\newblock {\em International Journal of Computer Vision}, 2005.

\bibitem{lee2017unsupervised}
Hsin-Ying Lee, Jia-Bin Huang, Maneesh Singh, and Ming-Hsuan Yang.
\newblock Unsupervised representation learning by sorting sequences.
\newblock In {\em ICCV}, 2017.

\bibitem{lee2018cleannet}
Kuang-Huei Lee, Xiaodong He, Lei Zhang, and Linjun Yang.
\newblock Cleannet: Transfer learning for scalable image classifier training
  with label noise.
\newblock In {\em CVPR}, 2018.

\bibitem{li2020learning}
Tianhao Li and Limin Wang.
\newblock Learning spatiotemporal features via video and text pair
  discrimination.
\newblock {\em arXiv:2001.05691}, 2020.

\bibitem{li2019temporal}
Yanghao Li, Sijie Song, Yuqi Li, and Jiaying Liu.
\newblock Temporal bilinear networks for video action recognition.
\newblock In {\em AAAI}, 2019.

\bibitem{ma2017less}
Shugao Ma, Sarah~Adel Bargal, Jianming Zhang, Leonid Sigal, and Stan Sclaroff.
\newblock Do less and achieve more: Training cnns for action recognition
  utilizing action images from the web.
\newblock {\em Pattern Recognition}, 2017.

\bibitem{mahajan2018exploring}
Dhruv Mahajan, Ross Girshick, Vignesh Ramanathan, Kaiming He, Manohar Paluri,
  Yixuan Li, Ashwin Bharambe, and Laurens van~der Maaten.
\newblock Exploring the limits of weakly supervised pretraining.
\newblock In {\em ECCV}, 2018.

\bibitem{miech2020end}
Antoine Miech, Jean-Baptiste Alayrac, Lucas Smaira, Ivan Laptev, Josef Sivic,
  and Andrew Zisserman.
\newblock End-to-end learning of visual representations from uncurated
  instructional videos.
\newblock In {\em CVPR}, 2020.

\bibitem{muller2020subclass}
Rafael M{\"u}ller, Simon Kornblith, and Geoffrey Hinton.
\newblock Subclass distillation.
\newblock {\em arXiv:2002.03936}, 2020.

\bibitem{patrick2020multi}
Mandela Patrick, Yuki~M Asano, Ruth Fong, Jo{\~a}o~F Henriques, Geoffrey Zweig,
  and Andrea Vedaldi.
\newblock Multi-modal self-supervision from generalized data transformations.
\newblock {\em arXiv:2003.04298}, 2020.

\bibitem{patrini2017making}
Giorgio Patrini, Alessandro Rozza, Aditya Krishna~Menon, Richard Nock, and
  Lizhen Qu.
\newblock Making deep neural networks robust to label noise: A loss correction
  approach.
\newblock In {\em CVPR}, 2017.

\bibitem{rupprecht2018learning}
Christian Rupprecht, Ansh Kapil, Nan Liu, Lamberto Ballan, and Federico
  Tombari.
\newblock Learning without prejudice: Avoiding bias in webly-supervised action
  recognition.
\newblock {\em CVIU}, 2018.

\bibitem{ryou2019anchor}
Serim Ryou, Seong-Gyun Jeong, and Pietro Perona.
\newblock Anchor loss: Modulating loss scale based on prediction difficulty.
\newblock In {\em ICCV}, 2019.

\bibitem{saxena2019data}
Shreyas Saxena, Oncel Tuzel, and Dennis DeCoste.
\newblock Data parameters: A new family of parameters for learning a
  differentiable curriculum.
\newblock In {\em NeurIPS}, 2019.

\bibitem{selvaraju2017grad}
Ramprasaath~R Selvaraju, Michael Cogswell, Abhishek Das, Ramakrishna Vedantam,
  Devi Parikh, and Dhruv Batra.
\newblock Grad-cam: Visual explanations from deep networks via gradient-based
  localization.
\newblock In {\em ICCV}, 2017.

\bibitem{sharma2020noiserank}
Karishma Sharma, Pinar Donmez, Enming Luo, Yan Liu, and I~Zeki Yalniz.
\newblock Noiserank: Unsupervised label noise reduction with dependence models.
\newblock {\em arXiv:2003.06729}, 2020.

\bibitem{simonyan2014two}
Karen Simonyan and Andrew Zisserman.
\newblock Two-stream convolutional networks for action recognition in videos.
\newblock In {\em NeurIPS}, 2014.

\bibitem{soomro2012ucf101}
Khurram Soomro, Amir~Roshan Zamir, and Mubarak Shah.
\newblock Ucf101: A dataset of 101 human actions classes from videos in the
  wild.
\newblock {\em arXiv:1212.0402}, 2012.

\bibitem{stroud2020learning}
Jonathan~C Stroud, David~A Ross, Chen Sun, Jia Deng, Rahul Sukthankar, and
  Cordelia Schmid.
\newblock Learning video representations from textual web supervision.
\newblock {\em arXiv:2007.14937}, 2020.

\bibitem{sun2019learning}
Chen Sun, Fabien Baradel, Kevin Murphy, and Cordelia Schmid.
\newblock Learning video representations using contrastive bidirectional
  transformer.
\newblock {\em arXiv:1906.05743}, 2019.

\bibitem{sun2015temporal}
Chen Sun, Sanketh Shetty, Rahul Sukthankar, and Ram Nevatia.
\newblock Temporal localization of fine-grained actions in videos by domain
  transfer from web images.
\newblock In {\em ACM international conference on Multimedia}, 2015.

\bibitem{tran2015learning}
Du Tran, Lubomir Bourdev, Rob Fergus, Lorenzo Torresani, and Manohar Paluri.
\newblock Learning spatiotemporal features with 3d convolutional networks.
\newblock In {\em ICCV}, 2015.

\bibitem{tran2018closer}
Du Tran, Heng Wang, Lorenzo Torresani, Jamie Ray, Yann LeCun, and Manohar
  Paluri.
\newblock A closer look at spatiotemporal convolutions for action recognition.
\newblock In {\em CVPR}, 2018.

\bibitem{wang2013action}
Heng Wang and Cordelia Schmid.
\newblock Action recognition with improved trajectories.
\newblock In {\em ICCV}, 2013.

\bibitem{wang2018non}
Xiaolong Wang, Ross Girshick, Abhinav Gupta, and Kaiming He.
\newblock Non-local neural networks.
\newblock In {\em CVPR}, 2018.

\bibitem{wu2018unsupervised}
Zhirong Wu, Yuanjun Xiong, Stella Yu, and Dahua Lin.
\newblock Unsupervised feature learning via non-parametric instance-level
  discrimination.
\newblock In {\em CVPR}, 2018.

\bibitem{xiao2015learning}
Tong Xiao, Tian Xia, Yi Yang, Chang Huang, and Xiaogang Wang.
\newblock Learning from massive noisy labeled data for image classification.
\newblock In {\em CVPR}, 2015.

\bibitem{xie2019self}
Qizhe Xie, Eduard Hovy, Minh-Thang Luong, and Quoc~V Le.
\newblock Self-training with noisy student improves imagenet classification.
\newblock In {\em CVPR}, 2020.

\bibitem{xie2018rethinking}
Saining Xie, Chen Sun, Jonathan Huang, Zhuowen Tu, and Kevin Murphy.
\newblock Rethinking spatiotemporal feature learning: Speed-accuracy trade-offs
  in video classification.
\newblock In {\em ECCV}, 2018.

\bibitem{yalniz2019billion}
I~Zeki Yalniz, Herv{\'e} J{\'e}gou, Kan Chen, Manohar Paluri, and Dhruv
  Mahajan.
\newblock Billion-scale semi-supervised learning for image classification.
\newblock {\em arXiv:1905.00546}, 2019.

\bibitem{yan2019clusterfit}
Xueting Yan, Ishan Misra, Abhinav Gupta, Deepti Ghadiyaram, and Dhruv Mahajan.
\newblock Clusterfit: Improving generalization of visual representations.
\newblock {\em arXiv:1912.03330}, 2019.

\bibitem{yeung2017learning}
Serena Yeung, Vignesh Ramanathan, Olga Russakovsky, Liyue Shen, Greg Mori, and
  Li Fei-Fei.
\newblock Learning to learn from noisy web videos.
\newblock In {\em CVPR}, 2017.

\bibitem{yue2018compact}
Kaiyu Yue, Ming Sun, Yuchen Yuan, Feng Zhou, Errui Ding, and Fuxin Xu.
\newblock Compact generalized non-local network.
\newblock In {\em NeurIPS}, 2018.

\end{thebibliography}
}

% \clearpage

\appendix
\section*{Appendix}
\addcontentsline{toc}{section}{Appendices}
\renewcommand{\thesubsection}{\Alph{subsection}}

\section{More Details of Network Structure and Implementation}
\textbf{Network structure.}
We describe more details of our backbone network R3D-50-G which is based on ResNet50~\cite{wang2018non}. An illustration of the backbone can be found in Table~\ref{tab:resnet50-g}. %We remove all max pooling in the temporal dimension.
A feature gating module~\cite{xie2018rethinking} is added after each residual block. Feature gating is a self attention mechanism that re-weights the channels based on context (e.g. the feature map averaged over time and space).

\begin{table}[!htb]
\centering
\resizebox{1.0\linewidth}{!}{
\begingroup
\setlength{\tabcolsep}{10pt}
\resizebox{1.0\linewidth}{!}{
\begin{tabular}{c | c | c }
Block  &   &  Output sizes $T\times S^2\times C$ \\
\hline
input & & $64\times224^2\times3$ \\
\hline
\multirow{2}{*}{$\text{conv}_1$} & $5\times7^2$ & \multirow{2}{*}{$64\times112^2\times64$} \\
& stride $1\times2^2$ & \\
\hline
\multirow{2}{*}{$\text{pool}_1$} & $1\times3^2$ & \multirow{2}{*}{$64\times56^2\times64$} \\
& stride $1\times2^2$ &  \\
\hline
$\text{res}_2$ & $\left[ \begin{array}{c} 3\times1^2 \\ 1\times3^2 \\ 1\times1^2 \end{array}\right] \times 3$ & $64\times56^2\times256$ \\
& feature gating & \\
\hline
$\text{res}_3$ & $\left[ \begin{array}{c} t_i\times1^2 \\ 1\times3^2 \\ 1\times1^2 \end{array}\right] \times 4$ & $64\times28^2\times512$ \\
& feature gating & \\
\hline
$\text{res}_4$ & $\left[ \begin{array}{c} t_i\times1^2 \\ 1\times3^2 \\ 1\times1^2 \end{array}\right] \times 6$ & $64\times14^2\times1024$ \\
& feature gating & \\
\hline
$\text{res}_5$ & $\left[ \begin{array}{c} t_i\times1^2 \\ 1\times3^2 \\ 1\times1^2 \end{array}\right] \times 3$ & $64\times7^2\times2048$ \\
& feature gating & \\
\end{tabular}
}
\endgroup
}
\centering
\vspace{0.2cm}
\caption{ ResNet50-G architecture used in our experiments. The kernel dimensions are $T \times S^2$ where $T$ is the temporal kernel size and $S$ is the spatial size. The strides are denoted as $\text{temporal stride}\times \text{spatial stride}^2$. For $res_3$, $res_4$, and $res_5$ blocks the temporal convolution only applies at every other cell. E.g., $t_i = 3$ when $i$ is an odd number and $t_i = 1$ when $i$ is even.}
\label{tab:resnet50-g}
\vspace{-0.5cm}
\end{table}

%As discussion, this section need to be finetuned from several aspects, citations, etc, to highlight our setting/hyperparams is fair to compared baselines

\textbf{Training.} For both pre-training and fine-tuning, we follow~\cite{wang2018non} to do random cropping on each input frame to get \(224\times224\) pixels from a scaled video whose shorter side is 256. We also perform random horizontal flipping and photometric augmentations such as randomly adjust brightness, contrast, hue and saturation. Synchronous stochastic gradient descent (SGD) is applied to train the model. Following \cite{wang2018non}, we use a momentum of 0.9 and weight decay of \(1\times10^{-7}\). Following \cite{chen2020simple,wu2018unsupervised}, we add a linear projection head during the pre-training. We adopt dropout \cite{hinton2012improving} and trainable BatchNorm \cite{ioffe2015batch} with the same setting as \cite{wang2018non}. We also apply anchor loss \cite{ryou2019anchor} for pre-training with emphasis on hard examples and find it sometimes can benefit the learning of SPLs. It brings around 0.3 improvements of top-1 accuracy on Mini-Kinetics-200 dataset when conducting SPL-B pre-training on WebK200-147K-V with \(6.7 \times 10^{5}\) clips. But the benefit is not obvious when conducting SPL-B pre-training on WebK200-285K-V set as we can achieve 85.3 of top-1 accuracy without using the anchor loss. It also does not benefit SPL-M and SPL-D on Mini-Kinetics-200 dataset.

\textbf{Inference.} We follow \cite{wang2018non} to perform inference on videos whose shorter side is re-scaled to 256. Following \cite{xie2018rethinking}, we sample 64 frames from the whole videos and conduct inference on Mini-Kinetics-200 dataset. A stride of 4 is applied during this sampling process.

% \subsection{Backbone}
% We describe details of our backbone network R3D-50-G which is based on ResNet50~\cite{wang2018non} with a few key modifications. First, we remove all max pooling in the temporal dimension. We find that applying temporal downsampling in any layer degrades the performance. Second, we add a feature gating module~\cite{xie2018rethinking} after each residual block. Feature gating is self attention mechanism that re-weights the channels based on context (e.g. the feature map averaged over time and space). We also explored adding feature gating modules after every residual cell which achieved similar results, so we decided to keep the former configuration given that it is more computationally efficient. An illustration of the backbone can be found in Table~\ref{tab:resnet50-g}. 

\begin{figure*}%[t]
\centering
\includegraphics[width=1.0\linewidth]{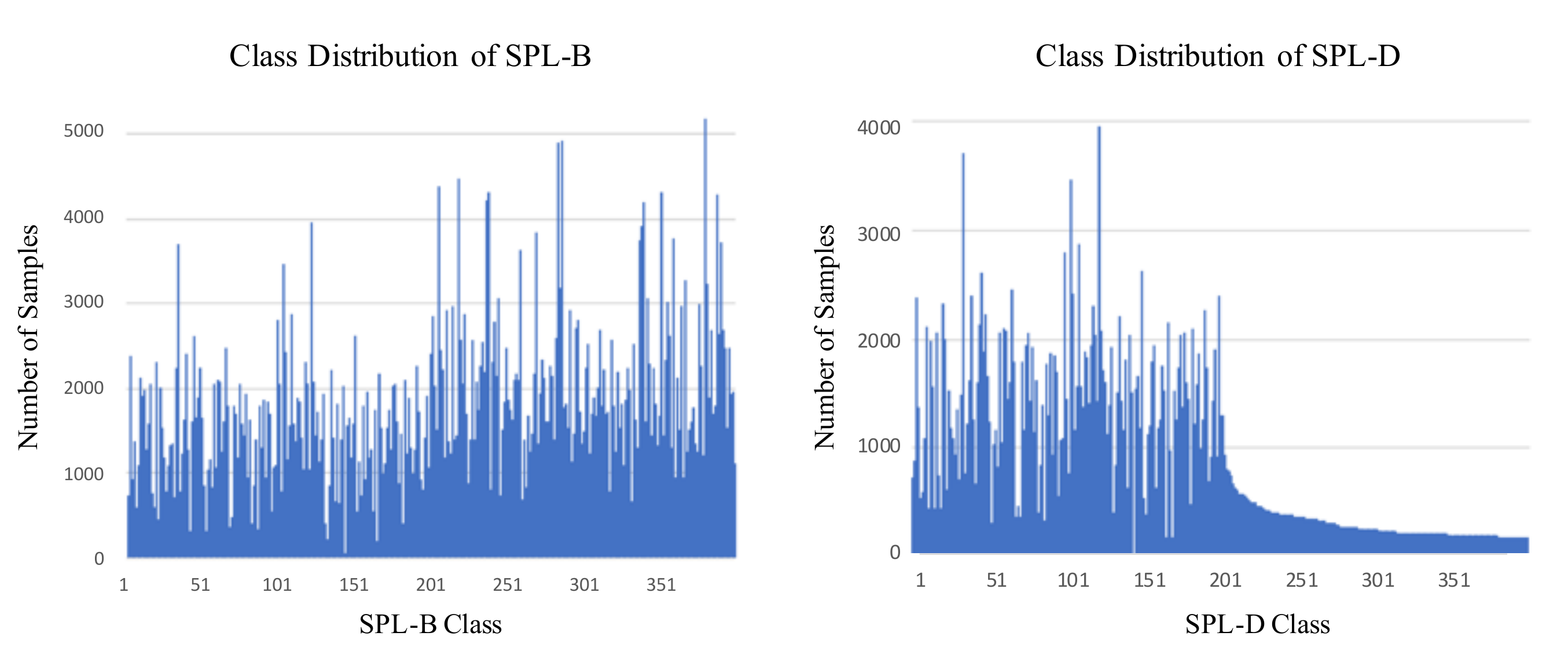} 
\caption{Class distribution of SPL on WebK200-285K-V set with \(6.7 \times 10^{5}\) clips. We show the cases of SPL-B (left) and SPL-D (right) with 400 SPL classes in total. The first 200 classes are in-diagonal SPL classes and rest 200 are off-diagonal classes. } 
\label{fig:sample_spl}
% \vspace{-3mm} %
\end{figure*}

\begin{figure*}%[t]
\centering
\includegraphics[width=0.9\linewidth]{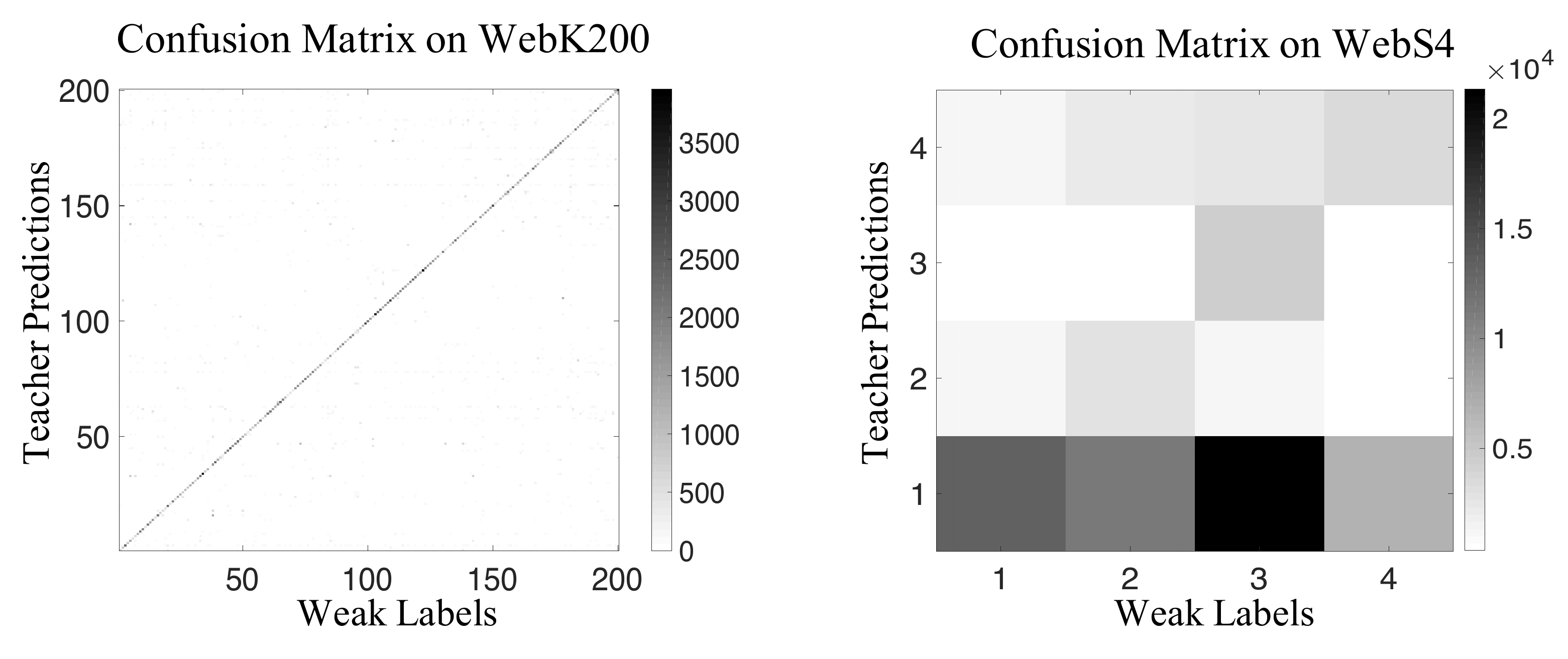} 
\caption{Confusion matrices on WebK200-285K-V (left, zoom in to better see off-diagonal elements) and WebS4-73K-C (right). They are web videos collected for target dataset Mini-Kinetics-200 and SoccerNet respectively. For WebS4, the four (1-4) classes are, Background, Yellow/Red Card, Goal, Substitution, respectively.} 
\label{fig:confusion_matrix}
% \vspace{-3mm}
\end{figure*}

\section{SPL and Confusion Matrix Statistics}
When we discuss the different variations of SPL in the main paper, we mentioned the long-tailed propriety for fine-grained SPL classes, especially for the original SPL, SPL-M and SPL-D. To verify this, we calculate the number of samples for each SPL class for WebK200-285K-V with \(6.7 \times 10^{5}\) clips. In Figure \ref{fig:sample_spl}, we show the cases of SPL-B (left) and SPL-D (right) with 400 SPL classes. The first 200 classes are in-diagonal SPL classes and rest 200 are off-diagonal SPL classes. We can find the long-tailed propriety exists in the off-diagonal SPL classes of SPL-D (SPL-M has a similar case). We also find the situation is much better for SPL-B, where the distribution of number of samples is more even for in-diagonal and off-diagonal SPL classes. 

In Figure \ref{fig:confusion_matrix}, we also show confusion matrices on WebK200 and WebKS4. They are web videos collected for target datasets Mini-Kinetics-200 and SoccerNet respectively. This aims to illuminate the difference of web videos collected for the common action recognition dataset and the fine-grained action recognition dataset. We can observe there exists high confusion between different classes on fine-grained one.

\section{More Details of Web Data Collection}
Here we include more details of web data collection described in Section 4.2 of the main paper. For data collection of WebK200 sets: WebK200-147K-V and WebK200-285K-V, we conduct duplicate checking on these two sets and web videos with addresses appearing in the validation set of Mini-Kinetics-200 are removed. 

We also list the full query terms used in collection of WebS4 sets for the target dataset SoccerNet. For the three foreground classes in SoccerNet: Goal, Yellow/Red Card, Substitution, we obtain the searching queries based on related terms from Wikipedia resulting in 9 kinds of queries in total. The queries for these three foreground classes are as follows: ``free kick goal'', ``corner kick goal'', ``own goal'', ``overhead kick goal'', ``last-minute goal'', ``ghost goal'', ``red card'', ``yellow card'', ``substitution of players''. For the Background class, it is hard to come up with queries for these background moments and people usually do not create video content to highlight them on the Internet.  Therefore, we use ``soccer full match'' as the query and randomly sample clips from the retrieved full match games. Because for entire soccer games, moments of foreground classes are very rare and sparse compared to the background events. Therefore, samples randomly sampled from the full game are very likely to belong to the Background class.

\begin{figure}[t]
\centering
\includegraphics[width=1.0\linewidth]{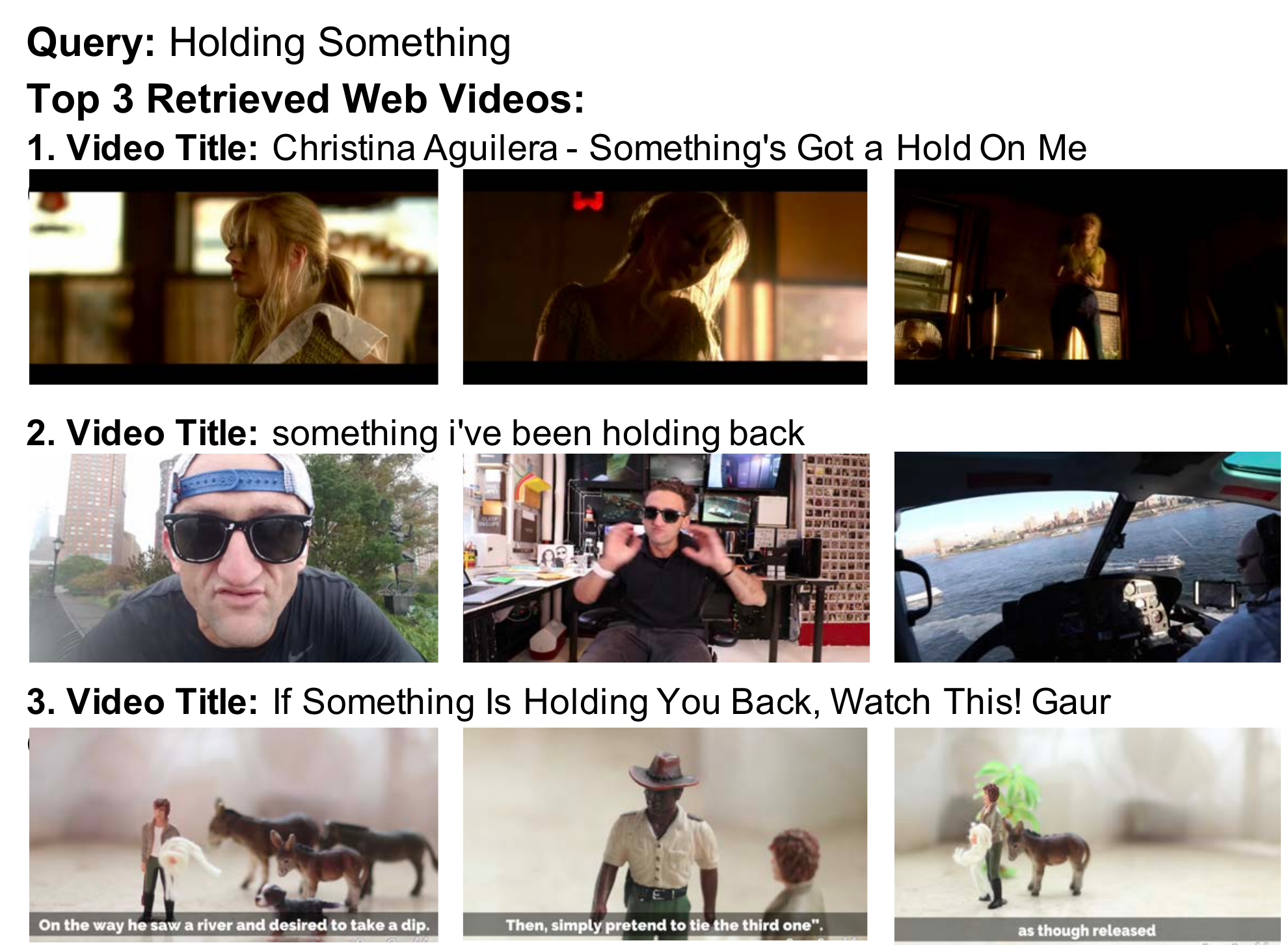} 
\caption{Top-3 retrieved videos obtained by taking the SS class ``Holding Something'' as the searching query. They are almost unrelated to the fine-grained motion defined in this class. } 
\label{fig:ss_1}
% \vspace{-3mm}
\end{figure} 

\begin{figure}[t]
\centering
\includegraphics[width=1.0\linewidth]{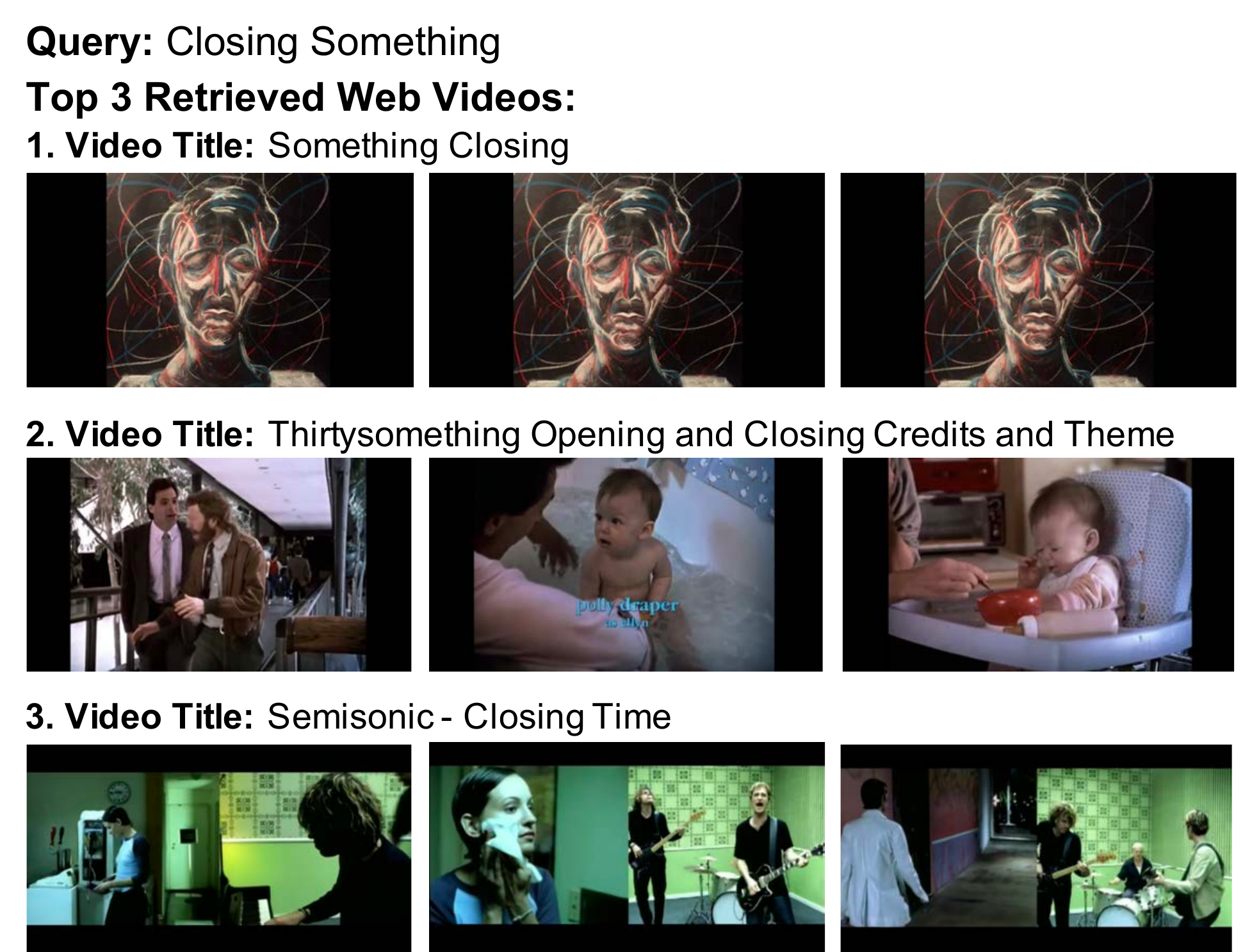} 
\caption{Top-3 retrieved videos obtained by taking the SS class ``Closing Something'' as the searching query. They are almost unrelated to the fine-grained motion defined in this class. } 
\label{fig:ss_2}
% \vspace{-3mm}
\end{figure}

\begin{figure}[t]
\centering
\includegraphics[width=1.0\linewidth]{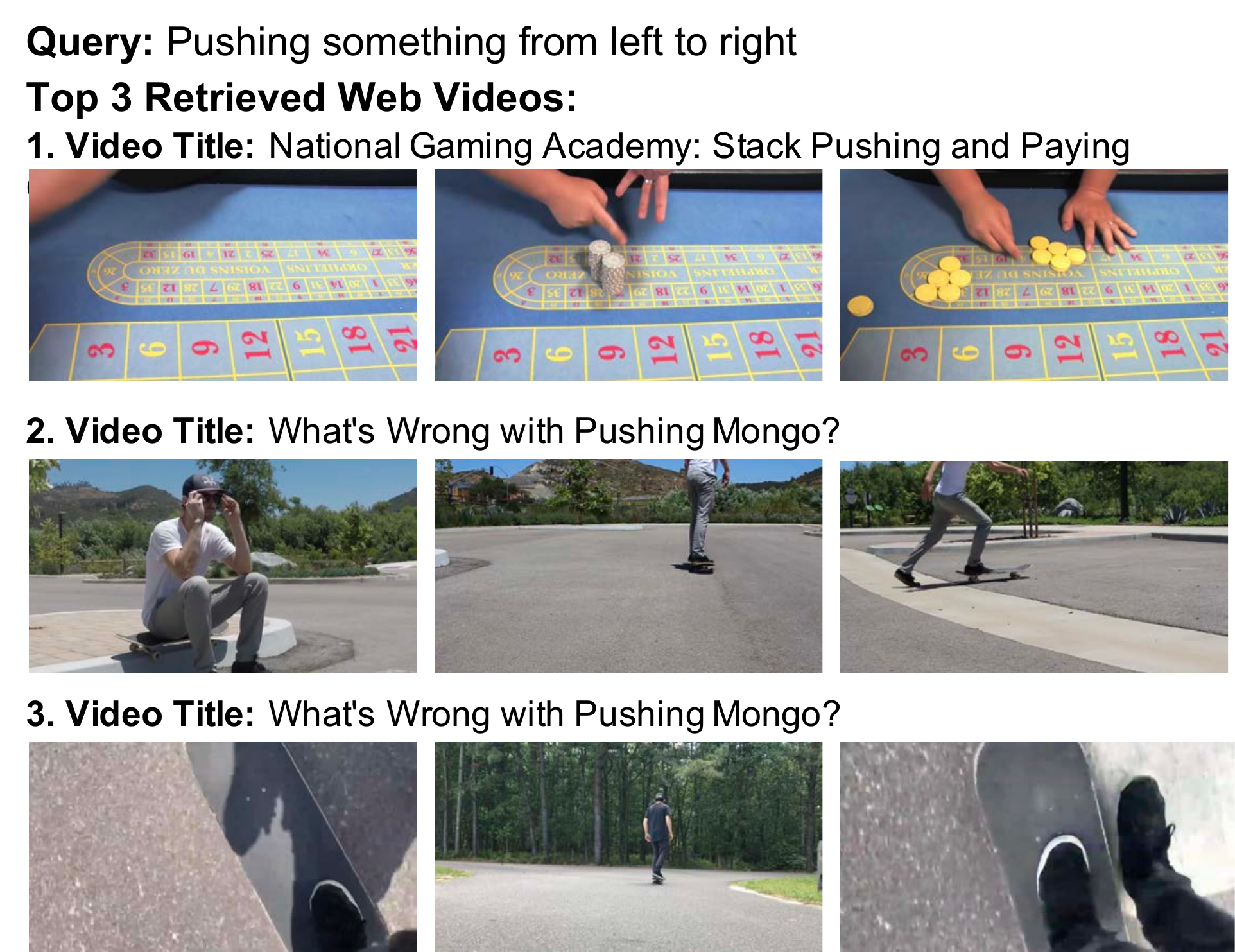} 
\caption{ Top-3 retrieved videos obtained by taking the SS class ``Pushing something from left to right'' as the searching query. They are almost unrelated to the fine-grained motion defined in this class. } 
\label{fig:ss_3}
% \vspace{-3mm}
\end{figure}

% \subsection{Research Challenges for Something-Something Dataset}
\section{Research Challenges for Dataset that Has Special Class Names}\label{sc:dataset_challenge}
% Reason and show some examples
One basic assumption for pre-training on web videos is that the search query is close related to the content of the retrieved videos so that these queries can be treated as weak labels to provide effective supervision during pre-training. There will be extra challenges when class names in the target dataset cannot be used as reliable search queries. For example, when conducting explorations on the fine-grained action datasets, we have considered the Something-Something (SS) \cite{goyal2017something} dataset initially but are blocked due to this problem. We find that the quality of the collected web videos are quite low when using action classes from SS dataset as queries. 
The reason may be that the class names in SS are uncommon in the metda data of web videos resulting in the retrieved videos are almost unrelated to the search queries.  
Several examples for the retrieved videos are shown in Figure \ref{fig:ss_1}, Figure \ref{fig:ss_2} and Figure \ref{fig:ss_3} by randomly selecting class names in SS as the search query. 
In such a case, there would be extreme label noise if treating the text query as the class label for these web videos, which does not satisfy the basic assumption for learning from weakly-labeled data. 
% This situation is also related to the collection process of SS, which is done by asking crowd workers to record videos themselves.
Addressing the acquisition of weakly-labeled web data with higher quality for this unique dataset could be a new research topic in this area.

% \section{Source Code}
% To encourage reproducibility of our method in an reasonable amount of efforts, we provide the source code of image classification experiments, which can be found in the supplementary material, \emph{SPL\_code.zip}. %checked: put the code into the supplementary material.

\end{document}